\documentclass[preprint,12pt]{elsarticle}
\usepackage{amsmath,amsfonts}
\usepackage{algorithmic}
\usepackage{algorithm}
\usepackage{array}
\usepackage[caption=false,font=normalsize,labelfont=sf,textfont=sf]{subfig}
\usepackage{textcomp}
\usepackage{stfloats}
\usepackage{url}
\usepackage{verbatim}
\usepackage{graphicx}
\usepackage{booktabs}
\usepackage{graphicx}
\usepackage{multirow}
\usepackage{pifont}
\usepackage{bbm}
\usepackage{enumitem}
\newcommand{\cmark}{\ding{51}}%
\newcommand{\xmark}{\ding{55}}%
\usepackage[colorlinks=true,linkcolor=red,urlcolor=green,citecolor=green]{hyperref}

\usepackage{amssymb}

\journal{Neural Networks}

\begin{document}
	
	\begin{frontmatter}
		
		
		\title{Reducing Semantic Ambiguity In Domain Adaptive Semantic Segmentation Via Probabilistic Prototypical Pixel Contrast}
		\author[au1]{Xiaoke Hao}
		\author[au1]{Shiyu Liu}
		\author[au1]{Chuanbo Feng}
		\author[au1]{Ye Zhu}
		\affiliation[au1]{organization={School of Artificial Intelligence, Hebei University of Technology},
			addressline={No. 5340, Xiping Road}, 
			city={Tianjin},
			postcode={300401}, 
			country={China}}
		
		
%
		
%
		
		\begin{abstract}
			Domain adaptation aims to reduce the model degradation on the target domain caused by the domain shift between the source and target domains. Although encouraging performance has been achieved by combining contrastive learning with the self-training paradigm, they suffer from ambiguous scenarios caused by scale, illumination, or overlapping when deploying deterministic embedding. To address these issues, we propose probabilistic prototypical pixel contrast (PPPC), a universal adaptation framework that models each pixel embedding as a probability via multivariate Gaussian distribution to fully exploit the uncertainty within them, eventually improving the representation quality of the model. In addition, we derive prototypes from probability estimation posterior probability estimation which helps to push the decision boundary away from the ambiguity points. Moreover, we employ an efficient method to compute similarity between distributions, eliminating the need for sampling and reparameterization, thereby significantly reducing computational overhead. Further, we dynamically select the ambiguous crops at the image level to enlarge the number of boundary points involved in contrastive learning, which benefits the establishment of precise distributions for each category. Extensive experimentation demonstrates that PPPC not only helps to address ambiguity at the pixel level, yielding discriminative representations but also achieves significant improvements in both  synthetic-to-real and day-to-night adaptation tasks. It surpasses the previous state-of-the-art (SOTA) by \mbox{+5.2\%} mIoU in the most challenging daytime-to-nighttime adaptation scenario, exhibiting stronger generalization on other unseen datasets. The code and models are available at \url{https://github.com/DarlingInTheSV/Probabilistic-Prototypical-Pixel-Contrast}.
		\end{abstract}
		



\end{frontmatter}
	
	
	\section{Introduction}
	Semantic segmentation is a dense prediction task at the pixel level, with the objective of assigning each pixel a corresponding label. It is a crucial step for enabling autonomous driving \cite{janai2020computer} and assisting medical analysis \cite{ronneberger2015u}. Though neural networks have achieved impressive performance on segmentation tasks, they require large-scale annotated datasets to stabilize the training process and prevent overfitting or inaccuracies in data statistics. It is particularly laborious to annotate each pixel for high-resolution images, for example, it takes 1.5 hours to annotate a single image of Cityscapes \cite{cordts2016cityscapes}, and for adverse weather conditions, it is even 3.3 hours. Therefore, adopting synthetic data rendered from game engines is a promising option. However, the model trained on synthetic images often experiences performance degradation on real data due to its sensitivity to domain shift. Additionally, the architecture that achieves higher performance in the supervised way does not necessarily mitigate this influence. For instance, DeepLabV3+ \cite{chen2018encoder} outperforms DeepLabV2 \cite{chen2017deeplab} in several supervised segmentation tasks, but it is more susceptible to the impact of domain gaps \cite{hoyer2022daformer}.
	
	To address this issue, unsupervised domain adaptation (UDA) methods have been proposed, which facilitate the adaptation of models trained on labeled source domains to unlabeled target domains. A lot of works delve into diminishing the domain shift in input \cite{dundar2020domain, yang2020fda}, feature \cite{hoffman2018cycada, tsai2018learning}, or output \cite{vu2019advent} space. A line of work employs adversarial learning, enforcing the encoder to generate domain-invariant features. However, alignment in a holistic way often leads to class-mismatching (e.g., style transfer and domain mix-up). Another line of work focuses on the self-training paradigm. They improve the self-training framework by refining the pseudo-label \cite{zhang2021prototypical}, using consistency regulation between different views of the same input \cite{araslanov2021self}, selecting reference pairs in the target domain \cite{wu2021dannet,sakaridis2019guided} or extracting fine-grain features \cite{wang2020classes}. Though these methods achieve remarkable improvement, rather than directly addressing the domain discrepancy, they implicitly mitigate the domain gap by separately optimizing performance in the source and target domain. Hence, several works combine adversarial training with self-training to help embedding space less vulnerable to domain discrepancy \cite{li2019bidirectional, li2023contrast}. Additionally, auxiliary tasks like self-supervised depth estimation \cite{vu2019dada,wang2021domain} and pixel-level contrast \cite{xie2023sepico} are also integrated into self-training. 
	
	Contrastive learning serves as a self-supervised strategy, enabling neural networks to learn to extract general features by optimizing label-agnostic tasks. The core of incorporating contrastive learning into UDA is to construct a coherent pixel embedding space across two domains. Therefore, most of the previous works deploy contrastive learning to explore the cross-domain pixel contrast, which attracts positive pairs and repels negative pairs.
	
\begin{figure*}[t]
		\centering
		\includegraphics[width=\textwidth]{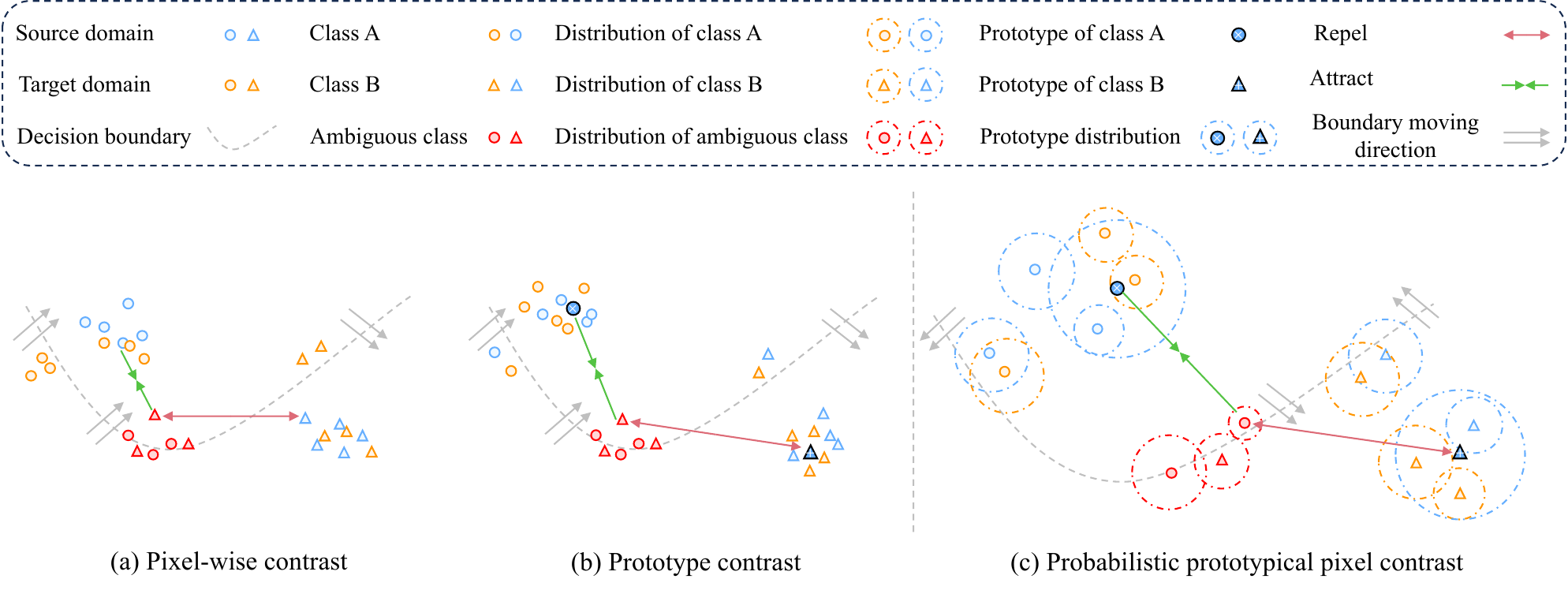}
		\caption{Illustration of the existing issues in self-training contrastive learning. (a) The decision boundary crosses the part of target pixel embeddings, leading to incorrect pseudo-label predictions. (b) The use of prototype contrast ignores the diversity of pixel embeddings, resulting in the decision boundary being unable to distinguish a few embeddings from both the source and target domains. (c) Our probabilistic prototypical pixel contrast not only better adjusts the decision boundaries but also addresses the issue of ambiguous classes that the previous two methods did not involve.}
		\label{fig1}
	\end{figure*}

	However, there are two main challenges for employing contrastive learning in UDA. One is the difficulty in obtaining a certain decision boundary. Due to a lack of supervision in the target domain, we utilize pseudo-label as the class label for pixel embedding. However, as shown in Fig. \ref{fig1}(a), some pixel embeddings have been incorrectly assigned labels, causing the decision boundary to move in the wrong direction. This bias becomes more serious with the training proceeding and eventually leads to a decrease in the model’s capability to discriminate similar classes in the target domain.  Meanwhile, it is infeasible to perform pixel-wise contrast if we want to reduce the computational overhead. To address these problems, prototype contrast is proposed. A prototype renders the overall appearance of a category by averaging the embeddings, as shown in Fig. \ref{fig1}(b). Although the computation is significantly reduced, it ignores the variety of pixel embedding of the category, thus still leading to part of pixel embedding incorrectly divided by decision boundary. The other challenge is that, whether using pixel-wise or prototype-based methods, deterministic embeddings fail to address the classification of ambiguous pixel embeddings. They can only determine classes based on the maximum value of the Softmax prediction, and in the worst case, each class has a similar prediction probability, much like a random guess. Therefore, they can’t provide an effective gradient for the loss function, thus leading the network into local optima.
	
	To address the above issues, we propose PPPC to deal with ambiguous embeddings in a more effective way by explicitly modeling uncertainty for each pixel embedding. Our idea is straightforward: for pixel embeddings with low uncertainty, we include them in the contrastive process. However, for pixel embeddings with high uncertainty that surpass the current model’s discriminative capability, we maintain their original positions, rather than randomly labeling them, leading to an incorrect optimization direction, as illustrated in Fig. \ref{fig1}(a) and Fig. \ref{fig1}(b). Therefore, in the worst-case scenario, the decision boundary will remain unchanged rather than moving in the wrong direction. Specifically, as shown in Fig. \ref{fig1}(c), in contrast to deterministic embeddings, we represent each pixel embedding as a multivariate Gaussian distribution, thus setting a dynamic uncertain range for them according to the model’s current discriminative capability. The prototype for each category is calculated based on the posterior distribution of all observations rather than being a simple average. Its position tends to align with high-confidence pixel embedding and the uncertainty range is affected by embeddings with a higher covariance. Therefore, our composed prototype can reduce the loss of diversity through a reasonable uncertainty range while ensuring an accurate position. By employing the expected likelihood kernel (ELK) as the measure of similarity between pixel and prototype in contrastive learning, we successfully direct the decision boundary towards the correct direction. Moreover, boundary points contain a lot of unexplored fuzzy information, enlarging the number of these points would be advantageous for better handling ambiguous classes. Hence, we propose ambiguity-guided cropping (AGC), which dynamically selects boundary points at the image level in a global manner based on prototype covariance. These prototypes serve as an effective approximation of the model’s current discriminative capability.
	
	The main contributions of this paper are summarized as follows:
	\begin{itemize}
		\item We propose a universal probabilistic adaptation framework, called PPPC, which can fully exploit the uncertainty information of each pixel embedding from the feature level to the data level. Consequently, it can significantly enhance segmentation performance in ambiguous scenarios.
		\item To facilitate efficiency, we carefully design each module and loss function. Specifically, we demonstrate that prototypes calculated from posterior probabilities can better account for the uncertainty of pixel embeddings compared to directly averaging. Additionally, we find that kernel functions are more stable and efficient than other divergence-based methods as a similarity metric under UDA settings. Finally, we introduce more boundary points into contrastive learning by approximating image norm through prototype variance, avoiding the manual setting of thresholds. As a result, PPPC significantly improves performance while only slightly increasing computation overhead and GPU memory footprint.
		\item Extensive experiments on three typical UDA tasks show that PPPC achieves superior performance on three wisely adopted UDA benchmarks. Particularly, we obtain mIoUs of  63.7\%, 58.7\%, and 50.6\% on benchmarks GTAV $\to$ Cityscapes, SYNTHIA $\to$ Cityscapes and Cityscapes $\to$ Dark Zurich. Especially on the most challenging day-to-night task, we improve the previous SOTA performance by +5.2\% mIoU.
	\end{itemize}
	
	\section{Related work}
	\subsection{Unsupervised Domain Adaptation Segmentation}
	Numerous approaches have been developed to address domain shifts in semantic segmentation, with the majority falling into two main categories: adversarial training and self-training. In adversarial training methods, typically two networks are employed: one network generates images, features, or segmentation maps, which may belong to either the source or target domain, while another network functions as a discriminator. The discriminator takes the generated output from the generator network as input and attempts to predict the domain of the input. Hoffman et al. \cite{hoffman2016fcns} first apply the adversarial training to UDA semantic segmentation. Other works \cite{peng2023diffusion,li2023contrast} use style transfer inputs, transferring the image style of the target domain onto the source domain images, thus bridging the domain gap by increasing domain confusion. Saito et al. \cite{saito2018maximum} propose to utilize task-specific classifiers to align distributions, avoiding generating target features near class boundaries. In \cite{vu2019advent}, to handle the problem that predictions on target images are less certain, resulting in noisy, high entropy output, the entropy of the pixel-wise predictions is incorporated in the adversarial loss. 
	
	In self-training, pseudo-labels are generated for the target domain using confidence thresholds \cite{mei2020instance,zhang2018collaborative} or pseudo-label prototypes \cite{pan2019transferrable,zhang2021prototypical}. To avoid training instabilities, several methods have been proposed. For example, Zhou et al. \cite{zhou2023adaptive} employ a novel knowledge distillation strategy to benefit the mutual learning of teacher and student networks. Tranheden et al. \cite{tranheden2021dacs} propose cross-domain mixed sampling, using mixed images from two domains along with their labels. In \cite{li2022class}, they employ a distribution alignment technique to enforce the consistency between the marginal distribution of clusters and pseudo labels to overcome the class imbalance problem that exists in pseudo labels. Hoyer et al. \cite{hoyer2022hrda} utilize a multi-resolution input fusion strategy that combines the strengths of high-resolution and low-resolution crops which provide fine details and long-range dependencies. Hoyer et al. \cite{hoyer2023mic} use a masking strategy to enforce consistency between predictions of masked target images. Most self-training studies encode images to deterministic embeddings and perform alignment or pseudo-label refinement, which is often problematic as they can't deal with inputs near the decision boundary that are ambiguous in both image and feature levels. We overcome this problem by fully exploring the potential of probabilistic embedding in self-training. Specifically, we not only leverage the uncertainty provided by probabilistic embedding to aid in addressing the alignment of boundary points in contrastive learning but also apply it to the dynamic selection of input crops.
	
	\subsection{Probabilistic embedding}
	The purpose of probabilistic embedding (PE) is to represent levels of specificity or uncertainty that traditional embeddings struggle to do. Vilnis et al.  \cite{vilnis2014word} propose directly working in probability distribution rather than applying Bayes’s rule to infer the latent distribution from observed data. Kingma et al. \cite{kingma2013auto} first propose to use MLP to model the mean and variance of the probability distribution. Oh et al. \cite{oh2018modeling} propose to design the embedding function to be stochastic and map the input to a certain probability distribution (e.g., Gaussian) instead of a single point. Li et al. \cite{li2021spherical} further extend the Gaussian to r-radius von Mises Fisher (vMF), and Kirchhof et al. \cite{kirchhof2022non} use a non-isotropic vMF. Shi et al. \cite{shi2019probabilistic} apply PE to face recognition, which not only improves the accuracy but shows a potential use in a risk-controlled system with estimated uncertainty. Recent literatures have delved deeper into PE, where Chun et al. \cite{chun2021probabilistic} apply it to cross-modal retrieval using an attention-based architecture to encode mean and variance, Neculai et al. \cite{neculai2022probabilistic} apply it to multimodal image retrieval with a probabilistic composition rule, and Xie et al. \cite{xie2023boosting} apply it in semi-supervised segmentation to deal with inaccurate pseudo-label. However, these methods are typically applied in sparse tasks like classification and retrieval, which encode the whole input into a single vector. Furthermore, they employ a complex Monte Carlo sampling strategy to compute the similarity of two distributions, which also requires additional reparametrization tricks to enable backpropagation during training. Instead, semantic segmentation is a dense prediction task, we treat each pixel embedding in the feature map as a probability distribution. We further demonstrate by experiment that measuring distribution similarity directly through mean and variance is more efficient than Monte Carlo sampling.
	
	\subsection{Contrastive learning}
	Contrastive learning,  first proposed in \cite{hadsell2006dimensionality}, involves the network minimizing the contrastive loss in the latent space. This is accomplished by pulling positive pairs closer while pushing negative pairs away, ultimately enhancing the network's discriminative ability. The pairs used for contrastive learning can be images \cite{he2020momentum}, pixels \cite{huang2022category}, and prototypes \cite{zhang2021prototypical}. For instance, Zhang et al. \cite{zhang2021prototypical} contrast the pixel representation with prototypes which are updated in a moving average way to learn a more compact embedding space. Jiang et al. \cite{jiang2022prototypical} further explore the inter-class alignment by adopting the class-centered distribution alignment for adaptation. Li et al. \cite{li2023contrast} bridge the domain gap at both pixel and feature levels via contrastive learning. Huang et al. \cite{huang2022category} increase the quantity of prototypes by establishing a domain-mixed memory bank that stores class-wise prototypes from both the source domain and target domain. Xie et al. \cite{xie2023sepico} pushe the number of contrast instances to infinite by approximating the true distribution of each semantic category in the source domain. However, they model distribution implicitly by using statistics from the source domain, we explicitly calculate the distribution for each instance involved in contrastive learning, including pixels and prototypes in both domains. Modeling the distribution of representations for each pixel in the feature map introduces only a little overhead, yet retains their original semantic information and preserves prototype diversity, which is hard to achieve even using the statistical distribution in deterministic embeddings. Xie et al. \cite{xie2023boosting} also employ the same idea in semi-supervised segmentation. However, it requires complex sampling strategies for valid samples, anchors, and negative samples. In contrast, our approach contrasts pixel embeddings only with prototypes, significantly reducing the computation.

	\section{Methodology}
	\subsection{Problem definition and overall framework}
	
		\begin{figure*}[!ht]
	\centering
	\includegraphics[width= \textwidth]{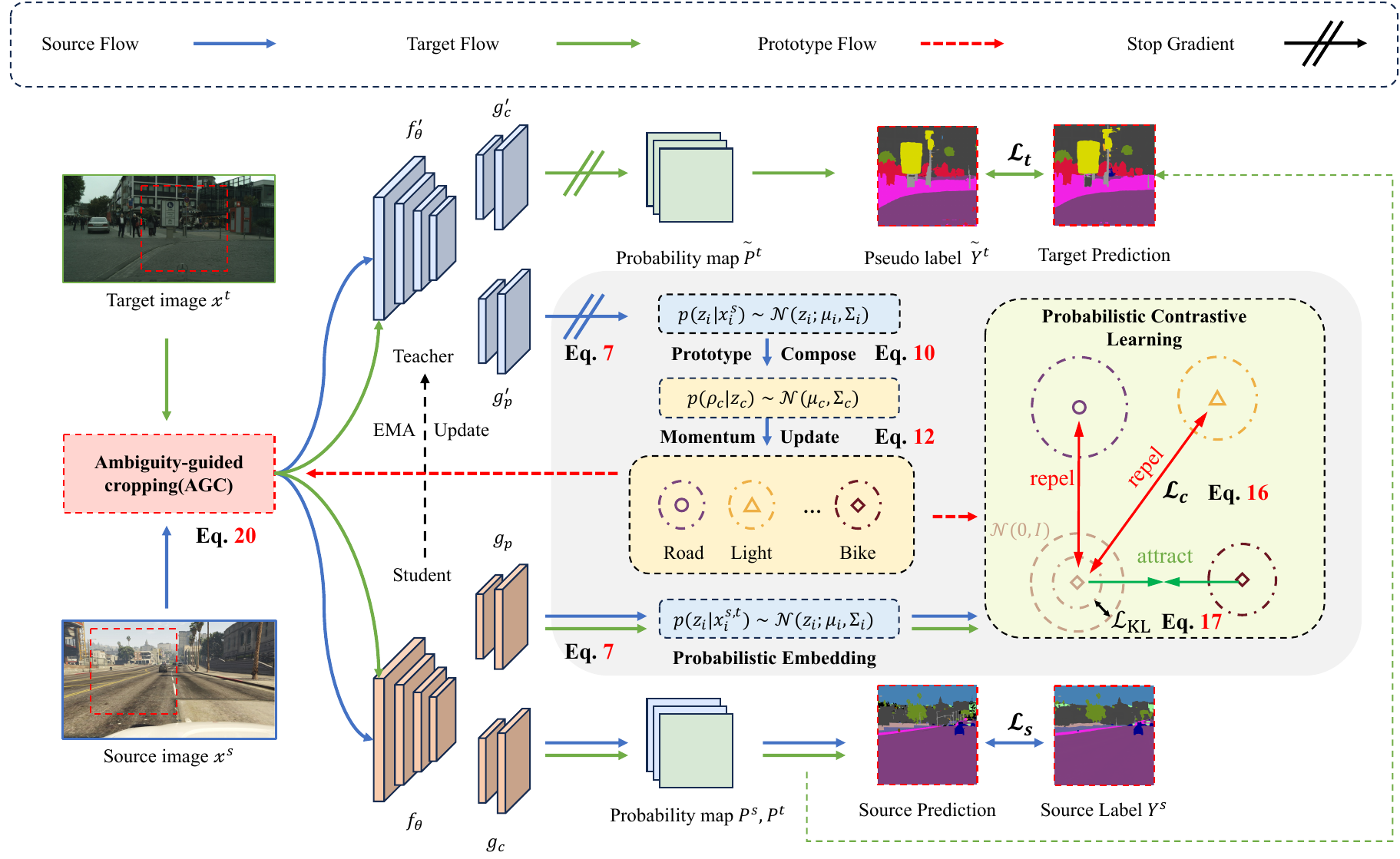}
	\caption{Overview of our framework. The model is trained with both the supervised segmentation loss $\mathcal{L}_s$ and the unsupervised adaptation loss $\mathcal{L}_t$. Specifically, the predictions for the source and target domain $P_s$, $P_t$ are provided by the student model, and the pseudo-labels $\tilde{Y}^t$ are generated by the teacher model. Furthermore, the distribution for each pixel embedding $p(z_i |x_i^{s,t})$ is modeled by the projection head of the student model. For prototypes, we first obtain pixel-wise probabilistic embeddings $p(z_i |x_i^s)$from the source image $x^s$ via the projection head $g_p^{\prime}$ within the teacher model. Then, we estimate each composed prototype $p(\rho_c |z_c)$ with these embeddings. Lastly, to stabilize the updating of prototypes we deploy the momentum update strategy. In the embedding space, apart from conducting distributional contrast, we also utilize an additional regularization term $\mathcal{L}_{KL}$ to prevent the covariance collapse to zero. Additionally, AGC uses the ambiguity provided by the online-updated prototypes to select ambiguous scenarios dynamically.}
	\label{fig2}
\end{figure*}
	
	Given the labeled source domain data $\mathcal{D}_{s}=\{(x_{i}^{s},y_{i}^{s})\}_{i=1}^{N_{s}}$  and unlabeled target domain data $\mathcal{D}_{t}=\{(x_{i}^{t})\}_{i=1}^{N_{t}}$, the objective of unsupervised domain adaptation of semantic segmentation is to train a model on $\mathcal{D}_{s}$ and $\mathcal{D}_{t}$ that can classify each pixel in a target domain image into one of $K$ predefined classes. We adopt the self-training framework as our basic architecture, as shown in Fig. \ref{fig2}. The self-training segmentation model has two networks sharing the same structure but different weight update strategies. The student networks have an encoder $f_{\theta}(\cdot)$, a segmentation head $g_c(\cdot)$, and an auxiliary projection head $g_p(\cdot)$. Similarly, the teacher networks are denoted as $f_\theta^{\prime}(\cdot)$, $g_c^{\prime}(\cdot)$ and $g_p^{\prime}(\cdot)$. 
	
	First, the student networks $g(f(\cdot))$ are trained on the labeled source domain in a supervised manner by minimizing the categorical cross-entropy between the model’s prediction $P_{i,c}^{s}$  and the ground truth label $ Y_{i,c}^{{s}}$. We formulate this problem as:
	\begin{equation}
		\label{eq1}
		\mathcal{L}_s=-\frac1{HW}\sum_{n=1}^{N_s}\sum_{i=1}^{H\times W}\sum_{c=1}^{K}Y_{n,i,c}^s\log P_{n,i,c}^s ,
	\end{equation}
	where $N_{s}$ is the total number of images in the source domain, $H$ and $W$ denote the height and the width of a source image, $K$ is the number of classes, $n$ is the index of image, $i$ is the pixel index of image, $c$ is the class index. $Y_{n,i,c}^s\in\{0,1\}$ is the one-hot representation of ground truth label for pixel $i$ in image $n$ about class $c$. $P_{i,c}^s\in\mathbb{R}$ is the predicted probability for pixel $i$ about class $c$, which is obtained by up-sampling the output of student networks $g_c(f(x_n^s))$. To better utilize the information in the unlabeled target domain, we generate reliable pseudo label $\tilde{Y_j^t}$ by teacher networks for a given target image: 
	\begin{equation}
		\label{eq2}
		\tilde{Y_j^t}=\underset{c}{\operatorname*{argmax}}\tilde{P_{j,c}^t}.
	\end{equation}
	Without supervised information, the generated pseudo label is prone to be noisy. Thus, we empirically set a threshold $\alpha$ to filter out low-confidence pixels:
	\begin{equation}
		\label{eq3}
		{M}(x_j^t)=\mathbbm{1}_{[max_c\widetilde{P}_{j,c}^t>\alpha]},j\in1,2,\cdots,H\times W.
	\end{equation}
	$M(\cdot)$ is sample mask over the pixels of a target image, $\mathbbm{1}_{[\cdot]}$ is the indicator function. Only the pixels whose predicted probability exceeds the threshold will participate in the re-training on the target domain:
	\begin{equation}
		\label{eq4}
		\mathcal{L}_t=-\frac1{HW}\sum_{n=1}^{N_t}\sum_{i=1}^{H\times W}\sum_{c=1}^{K}M_{n,i}^t\tilde{Y}_{n,i,c}^t\mathrm{log~}P_{n,i,c}^t.
	\end{equation}
	The parameters of student networks are optimized via gradient descent, while the teacher network parameters are updated by the Exponential Moving Average (EMA) of the student network parameters. The contrastive loss is computed based on the output of the projection head $q=g_p(f(x^{s,t}))$:
	\begin{equation}
		\label{eq5}
		\mathcal{L}_c=\mathbb{E}_{q,k^+,k^-}[-\log\frac{e^{s(q,k^+)/\tau}}{e^{s(q,k^+)/\tau}+\sum_{n=1}^Ne^{s(q,k^-)/\tau}}],
	\end{equation}
	where $k^+$ and $k^-$ are positive and negative pairs with respect to query $q$, they can be either pixel representation or prototype of class.  $s(\cdot, \cdot)$ denotes the similarity measurement between two representations, $\tau$ denotes the temperature. Intuitively, the goal of contrastive learning is to minimize the distance when pixels belong to the same category while maximizing the distance otherwise. By leveraging contrastive learning as an auxiliary task, the representations produced by the backbone encoder $f_{\theta}(\cdot)$ , used for downstream semantic segmentation task, are refined to have clearer decision boundaries among each class. The overall loss for UDA semantic segmentation is formulated as:
	\begin{equation}
		\label{eq6}
		\mathcal{L}=\mathcal{L}_s+\lambda_t\mathcal{L}_t+\lambda_c\mathcal{L}_c,
	\end{equation}
	where $\lambda_t$ and $\lambda_c$ are hyper-parameters controlling the strength of corresponding loss.

	As previously mentioned, conventional deterministic embedding is susceptible to the influence of unreliable pseudo labels and ambiguous classes, particularly during the initial stages of training. This can result in the model displaying a bias towards the majority and easily distinguishable classes, while allocating less attention to the minority and ambiguous classes, such as train, bus, bike, and motorbike. Once this bias is established, the model cannot rectify it solely through the supervised signals from the source domain, as there exists a gap between the source and target domain. If this trend persists, the decision boundary between each class will become indistinct, resulting in embedding points being situated closer to the boundary rather than being maximally separated. Consequently, to overcome this issue, we promote to use of probabilistic embedding which has more statistical information compared to deterministic embedding. Moreover, this probabilistic mechanism could be viewed as utilizing distributional uncertainty to expand the embedding space for hard, confusing classes, which is hard to achieve by using deterministic embedding.

	\subsection{Probabilistic embedding}
	The prevailing contrastive learning method focuses on deterministic embeddings, whose similarities are easily measured by dot production. However, external factors like illumination, overlapping, and object size may confuse the model thus could lead to ambiguous embeddings. Although we can’t eliminate these factors, we can explicitly model each pixel as a distribution in the embedding space, which allows the model to express uncertainty when the input is ambiguous. We parametrize the embedding of pixels as a multivariate normal distribution with mean vectors and covariance matrices in $\mathbb{R}^d$:
	\begin{equation}
		\label{eq7}
		p(z_i|x_i)=\mathcal{N}(z_i;\mu_i,\Sigma_i),
	\end{equation}
	where $\mu_i$ denotes the mean of the embedding pixel $i$, $\Sigma_i$ denotes the covariance, we only consider a diagonal covariance matrix to reduce the complexity (i.e., we model the covariance matrix of pixel embeddings and prototypes as corresponding diagonal scalars). $\mu_i$ and $\Sigma_i$  are predicted by two fully connected layers in projection head $g_p(f(x_i))$ respectively, the details are elaborated in the experiment section \ref{detail}.
	
	\subsection{Composed probabilistic prototype}
	Let $Z_{c}=\{z_{1},z_{2},\cdots,z_{k}\},k\in n_{c}$ be the set of $k$ embeddings belonging to the same class $c$. The objective of aggregating all pixel embeddings that share the same class label is to determine a probability distribution $p(\rho_c|Z_c)$that maximally unifies all the individual distributions $p(\rho_c|z_c)\sim\mathcal{N}(\mu_c,\Sigma_c)$. To be noticed, we do not directly estimate prototypes from pixel embeddings as done in \cite{xie2023sepico}, but instead, we utilize the posterior probabilities constructed from the likelihood function, similar to \cite{neculai2022probabilistic}. Since each pixel is modeled as a distribution, to maintain inter-class diversity as much as possible, we cannot simply average those observations. Additionally, due to the insufficient discriminative capability of the model, we must also account for their uncertainty. Therefore, we compute the posterior probabilities of $n$ observations, and the mean of the prototypes will be constrained by the covariances of individual observations. A simple way to compose this prototype after $n_c$ observed embeddings is making a product of $n_c$ Gaussian probability density functions (PDFs). Formally, the multivariate Gaussian PDF of pixel embedding Eq. \eqref{eq7} can be written as:
	\begin{equation}
		\label{eq8}
		p(z)=\exp[\zeta+\eta^Tz-\frac12{z}^T\Lambda z],
	\end{equation}
	where 
	\begin{equation*}
		\Lambda=\Sigma^{-1},\eta=\Sigma^{-1}\mu  \mathrm{~and~}\zeta=-\frac12(d\mathrm{log}2\pi-log|\Lambda|+\eta^T\Lambda^{-1}\eta),
	\end{equation*}
	where $d$ is the dimensionality of $z$, $\mu$ is the $d$-dimensional mean vector, and $\Sigma$ is the $d$-by-$d$ dimensional covariance matrix. The posterior distribution of prototype can be written as:
	\begin{align}
		\label{eq9}
		p(\rho_c) &=\prod_{i=1}^{n_c}p(z_i)\notag\\
		&=\exp[\zeta_{i=1...n_c}+(\sum_{i=1}^{n_c}\eta_i)^Tz-\frac12z^T(\sum_{i=1}^{n_c}\Lambda_i)z] \notag\\
		&=\exp(\zeta_{i=1...n_c}-\zeta_{n_c})\exp[\zeta_{n_c}+\eta_{n_c}^Tx-\frac12z^T\Lambda_{n_c}z],
	\end{align}
	where
	\begin{equation*}
		\Lambda_{n_c}=\sum_{i=1}^{n_c}\Lambda_i\mathrm{~and~}\eta_{n_c}=\sum_{i=1}^{n_c}\eta_i.
	\end{equation*}
	Comparing Eq. \eqref{eq8} to Eq. \eqref{eq9}, it can be seen that the new composite prototype distribution is a constant scaled Gaussian with a mean vector and covariance matrix given by:
	\begin{equation}
		\label{eq10}
		\mu_\rho=\frac{\sum_{i=1}^{n_c}{\Sigma}_i^{-1}{\mu}_i}{{\Sigma}_\rho^{-1}}\mathrm{~and~} \Sigma_\rho^{-1}=\sum_{i=1}^{n_c}\Sigma_i^{-1}.
	\end{equation}
	To make sure our composite prototypes are representative and reliable, we use the teacher network to extract the mean and covariance for each pixel embedding involved in the computation of Eq. \eqref{eq10}:
	\begin{equation}
		\label{eq11}
		\begin{aligned}
			{\mu}_i&=g_p^{\prime}(f^{\prime}(x^{s}_i))\\
			{\Sigma}_i&=g_p^{\prime}(f^{\prime}(x^{s}_i)).
		\end{aligned}
	\end{equation}
	To be noticed, we only calculate prototypes on the source domain which has label information rather than on the target domain which causes disturbance during self-training. Unlike some existing methods using a memory bank to store the prototypes at each training stage, we conjecture that probabilistic embedding is capable enough to deal with unstable changes of prototypes in embedding space during the initial training stage with a high learning rate. We only adopt an EMA update strategy to limit the prototypes to be similar in a few iterations in case of random error.
	\begin{equation}
		\label{eq12}
		\begin{aligned}
			{\mu}^{\prime}&\leftarrow\beta{\mu}^{\prime}+(1-\beta){\mu}\\
			{\Sigma}^{\prime}&\leftarrow\beta{\Sigma}^{\prime}+(1-\beta){\Sigma},
		\end{aligned}
	\end{equation}
	where ${\mu}^{\prime}$ and ${\Sigma}^{\prime}$ are mean and covariance from the current training loop, ${\mu}$ and ${\Sigma}$ are from the previous iteration.
	
	\subsection{Distribution to distribution distance}
	Since pixel embeddings and prototypes are not modeled as points but as distributions, conventional Euclidean distance cannot measure the similarity between two distributions.  To address this issue, we leverage a kernel function to measure the similarity between two distributions, which combines a weighted $l_2$ distance with a logarithmic regularization term, taking into account the corresponding reliability. Probability Product Kernels (PPK) are a family of metrics to compare two distributions by the product of their PDFs. Given two $d$-dimensional Gaussian distribution $p\sim\mathcal{N}(\mu_1,\Sigma)$ and $p\sim\mathcal{N}(\mu_2,\Sigma^{\prime})$, the PPK can be denoted as:
	\begin{equation}
		\label{eq13}
		PPK(p,q)=\int p(z)^\rho q(z)^\rho dz.
	\end{equation}
	Combining with Eq. \eqref{eq8}, PPK for two distributions can be written as:
	\begin{multline}
		\label{eq14}
		PPK(p,q)=(2\pi)^{(1-2\rho)D/2}\rho^{-D/2}\big|\Sigma^\dagger\big|^{1/2}|\Sigma|^{-\rho/2}|\Sigma^{\prime}|^{-\rho/2}\\\
		\exp\left(-\frac\rho2\big(\mu^T\Sigma^{-1}\mu+{\mu^{\prime}}^{-1}-{\mu^\dagger}^T\Sigma^\dagger{\mu^\dagger}\big)\right),
	\end{multline}
	where 
	\begin{equation*}
		{\Sigma}^\dagger=\left({\Sigma}^{-1}+{\Sigma}^{\prime{-1}}\right)^{-1}\mathrm{~and~}{\mu}^\dagger={\Sigma}^{-1}{\mu}+{\Sigma}^{\prime{-1}}{\mu}^{\prime}.
	\end{equation*}
	Here, we consider the PPK with $\rho=1$, it is called the ELK. In practice, we compute them with logarithm to ensure numerical stability as follows:
	\begin{align}
		\label{eq15}
		ELK(p,q)=-\frac{D}{2}\mathrm{log}2\pi-\frac{1}{2}\log({\Sigma}_1+{\Sigma}_2)+\frac{(\mu_1-\mu_2)^2}{{\Sigma}_1+{\Sigma}_2}.
	\end{align}
	In essence, ELK provides a tractable probabilistic formulation to measure the similarity between two distributions. Other metrics like divergence and cosine similarity are less effective and efficient than ours, as we will illustrate in the  experiment section \ref{distribution}. 
	
	\subsection{Probabilistic contrastive learning}
	Similar to conventional deterministic contrastive learning, our objective is imposed to pull the distribution of the source-composed prototype embeddings and both the source and target image pixel embeddings closer, while pushing away the distributions of negative pairs. Compared to the most used cosine similarity $s(z_{1},\rho)=\frac{z_{1}\cdot\rho}{\|z_{1}\|\|\rho\|}$ , which only leverages angles between them, leaving class-specific distribution covariance ignored, we propose to use distribution similarity metrics mentioned in Eq. \eqref{eq15}, accounting for either mean and covariance.
	\begin{equation}
		\label{eq16}
		\mathcal{L}_c=\mathbb{E}_{q,\mu^+,\mu^-}[-\log\frac{e^{\kappa(q,\mu^+)/\tau}}{e^{\kappa(q,\mu^+)/\tau}+\sum_{n=1}^Ne^{\kappa(q,\mu^-)/\tau}}],
	\end{equation}
	where $q$ denotes query pixel embedding, $\mu^+$ and $\mu^-$  denote positive and negative prototype embedding with respect to query $q$. In practice, we extract pixel embeddings from both the source and target domains as $q$, and based on the corresponding labels or pseudo-labels, we consider prototypes belonging to the same class as $q$  as $\mu^+$, while those belonging to different classes are considered as $\mu^-$. $\kappa(\cdot, \cdot)$ denotes kernel function which measures the similarity between two distributions. It is obvious that the $\kappa(\cdot, \cdot)$ is left unbounded if the covariance $\Sigma$ converges to zero. To make covariance a valid representation, we employ the additional KL regularization term between the pixel embedding distribution and the unit Gaussian prior $\mathcal{N}(0,I)$ to prevent the covariance collapse to zero:
	\begin{equation}
		\label{eq17}
		\mathcal{L}_{\mathrm{KL}}=\mathrm{KL}(p(z_i|x^{x,t}||\mathcal{N}(0,I)).
	\end{equation}
	Therefore, the overall training objective is formulated as:
	\begin{equation}
		\label{eq18}
		{\mathcal L}={\mathcal L}_{s}+\lambda_{t}{\mathcal L}_{t}+\lambda_{c}{\mathcal L}_{c}+\lambda_{KL}{\mathcal L}_{KL},
	\end{equation}
	where $\lambda_{t}$, $\lambda_{c}$, $\lambda_{KL}$ control the trade-off of the corresponding loss. 
	
	\subsection{Ambiguity-guided cropping}
	\begin{figure}[!t]
		\centering
		\includegraphics[width=0.6 \linewidth]{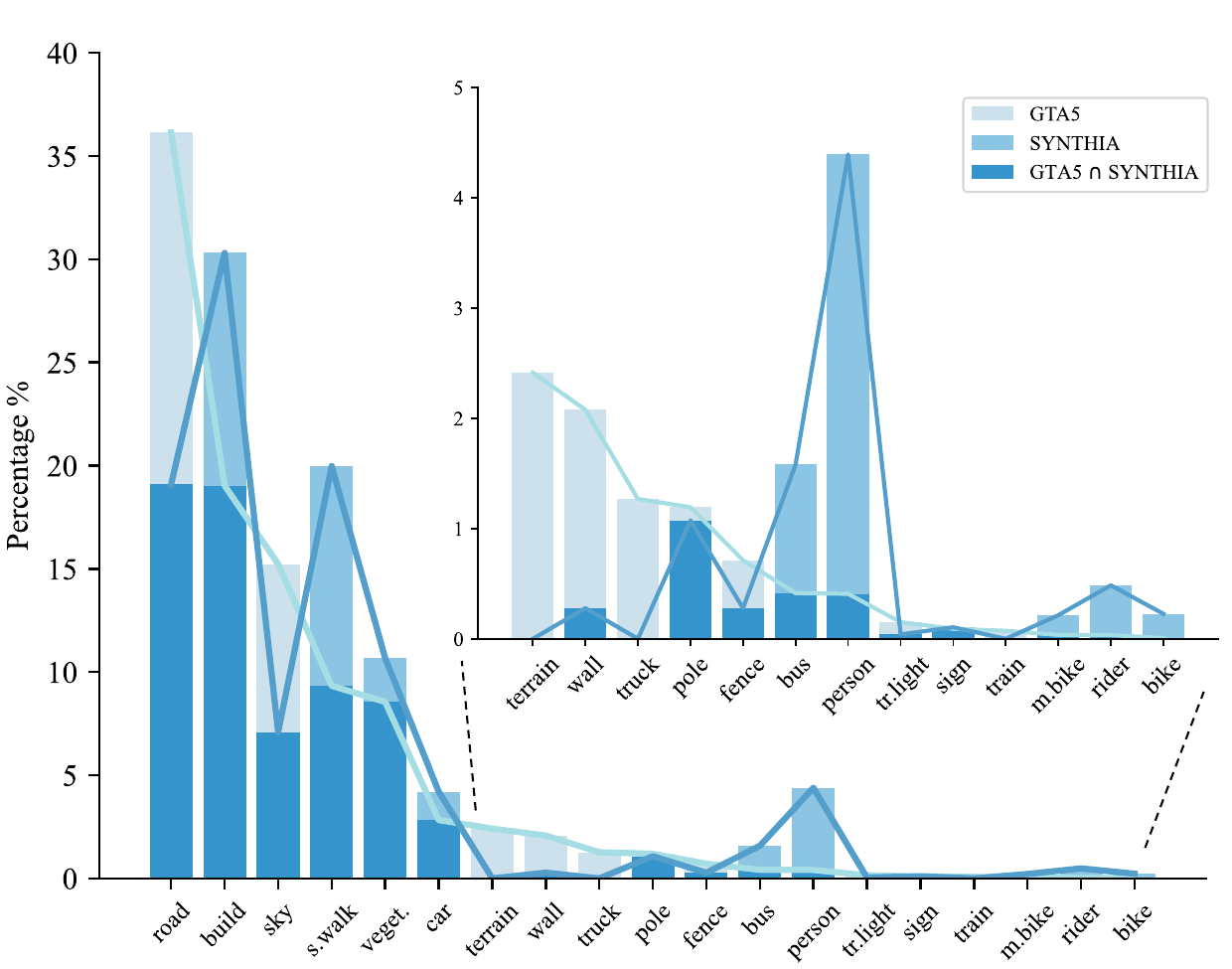}
		\caption{The Class distribution of the Cityscapes and SYNTHIA datasets.}
		\label{fig3}
	\end{figure}
	Class balance is hard to achieve for almost all of the dataset except with careful splitting. An imbalanced dataset often leads to overfitting to majority classes in the source domain. Rare class sampling (RCS) \cite{hoyer2022daformer} is one of the ways to solve the probslem, however, there are no labels available in the target domain. Class-balanced cropping (CBC) \cite{xie2023sepico} proposes to utilize generated target pseudo labels to crop image regions that jointly promote class balance in pixel number and diversity of internal categories. Though this strategy reduces the influence of imbalance, the discriminative capability improves less significantly because the model gains little on a balanced crop which only considers the numerical class balance. We suggest that in the latter stages of training, the difficulty of classes has a greater impact on the model's performance than class balance. To handle this issue, we propose Ambiguity-guided cropping (AGC) which utilizes the class-wise ambiguity provided by the prototypes rather than pixel embeddings, as the sample size for each class varies greatly, leading to inaccurate estimation. More specifically, we first calculate the ambiguity of each class by:
	\begin{equation}
		\label{eq19}
		a_i=\operatorname{softmax}({\Sigma}_i/\tau),
	\end{equation}
	where ${\Sigma}_i$ denotes prototype covariance of class $i$, $\tau$ denotes the temperature. As shown in Fig. \ref{fig3}, there are significant differences in the occurrence frequencies of different classes. Since the labels in the target domain are not available, to mitigate the impact of this imbalance on the estimation of crop uncertainty, we assume that the class distribution in the target domain is similar to that in the source domain. It can be observed that although GTAV and SYNTHIA have inconsistent percentages in a very few classes, they both exhibit a long-tailed distribution overall. Specifically, we only select the top-k classes with similar frequency levels as candidate classes, reducing the impact of class imbalance on crop ambiguity estimation. We calculate the overall score for each random select crop by:
	\begin{equation}
		\label{eq20}
		score_j=\sum_{i=0}^{n-1}\mathbbm{1}_{i\in S_{\mathrm{top\_k}}}a_in_{ij},
	\end{equation}
	where $j$ is the crop index, $i$ denotes the class, $S_{\mathrm{top\_k}}$  denotes the set only contains top-k ambiguity classes, $\mathbbm{1}_{[\cdot]}$ is an indicator function, $n_{ij}$ denotes pixel number for  class $i$ in crop $j$. In practice, given a target image with the original resolution, we random crop it for $N$ times, and calculate the score for each crop, the crop with the highest score is selected as the input image for the network. From Eq. \eqref{eq19} and Eq. \eqref{eq20}, we can see that RCS is a special case of AGC when at the initial stage of training, all ${\Sigma}_i$ have the same value, however, with the update of the prototypes, the crops are prone to involve more ambiguous scenarios, which better helps our PPPC in adjusting the decision boundary. Moreover, our AGC can be regarded as introducing image norm through prototypes as a selection criterion to boost the model’s discriminative capability without a forward propagation. This is in line with the finding that CNNs encode the amount of visible class discriminative features in the norm of the embedding \cite{scott2021mises}.
	
	\section{EXPERIMENT}
	
	\subsection{Datasets and Evaluation Metrics} \label{detail}
\begin{figure}[h]
		\centering
		\includegraphics[width=0.8 \linewidth]{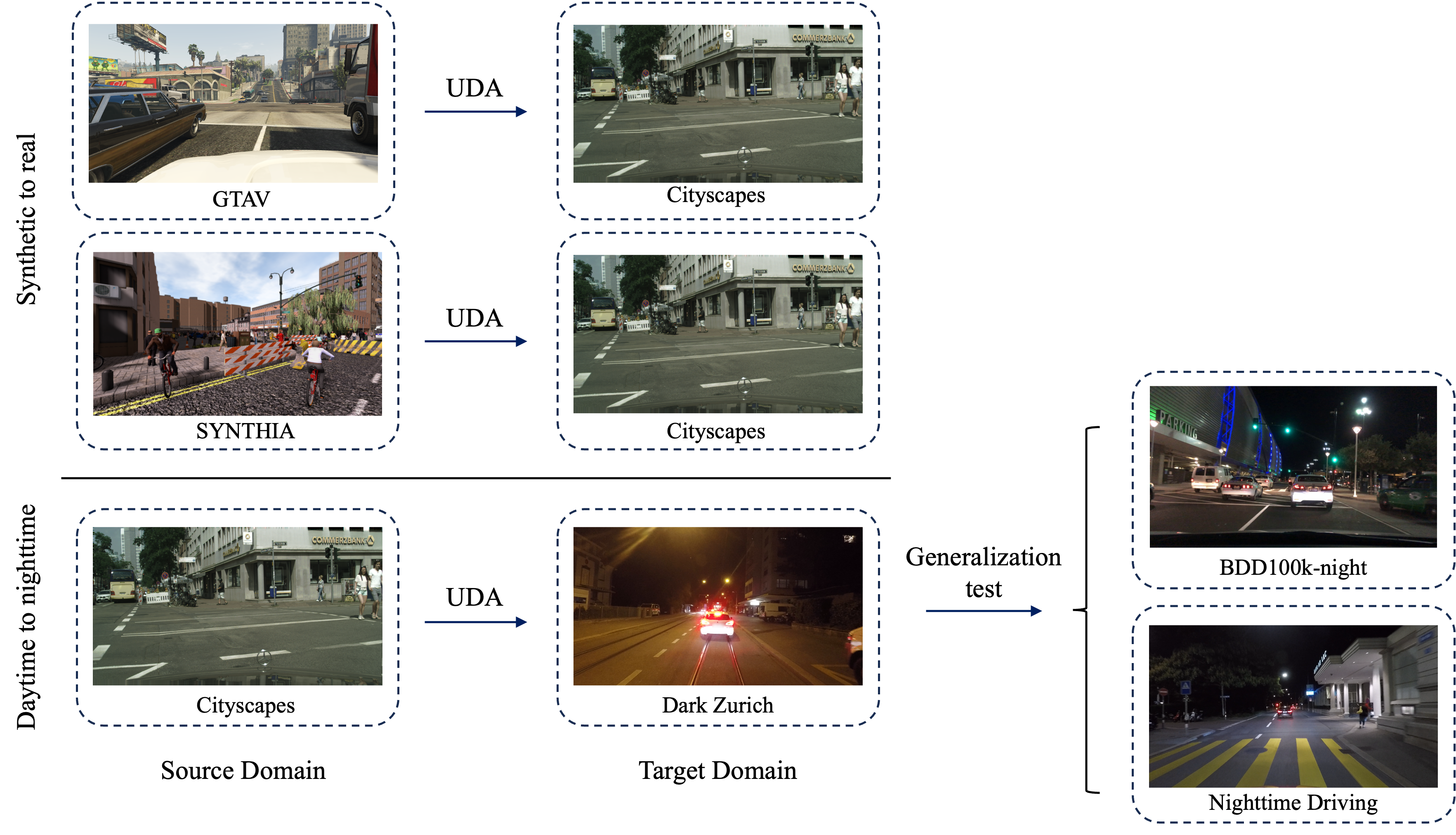}
		\caption{Overview of all our tasks. Our tasks include two synthetic-to-real tasks and one daytime-to-nighttime task. Additionally, we conduct extra generalization tests on the daytime-to-nighttime task.}
		\label{fig4}
			\end{figure}

	We evaluate our method on three cross-domain semantic segmentation tasks: GTAV $\to$ Cityscapes, SYNTHIA $\to$ Cityscapes and Cityscapes $\to$ Dark Zurich. Two synthetic datasets GTAV and SYNTHIA are served as labeled source domains, and Cityscapes is served as unlabeled target domain. For day-to-night adaptation, Cityscapes is utilized as source domain, Dark Zurich is served as the target domain. The overview of all tasks is shown in Fig. \ref{fig4}.
	\begin{enumerate}[wide]
		\item GTAV \cite{richter2016playing} is a composite dataset sharing 19 classes with Cityscapes, which contains 24,966 images with a resolution of 1914$\times$1052. 
		\item SYNTHIA \cite{ros2016synthia} is a synthetic urban scene dataset, we select its subset SYNTHIA-RAND-CITYSCAPES which has 16 common semantic annotations with Cityscapes and contains 9,400 training images with a resolution of 1280$\times$760. 
		\item Cityscapes \cite{cordts2016cityscapes} is a real urban scenes dataset taken from cities in Europe, we use 2,975 annotated images at the training stage, and 500 validation images during evaluation, all the images have a resolution of 2048$\times$1024. 
		\item Dark Zurich \cite{sakaridis2019guided} is a large dataset with urban driving scenes. It consists of 2,416 nighttime images, 2,920 twilight images, and 3,041 day-time images, with a resolution of 1920$\times$1080. Following \cite{xie2023sepico}, we use 2,416 day-night image pairs as target training data and another 151 test images as target test data. We obtain the mIoU result by submitting the segmentation predictions to the online evaluation website\footnote{https://competitions.codalab.org/competitions/23553}.
		\item BDD100k-night \cite{yu2020bdd100k}. To verify the generalization of our model, we directly use our model trained from Cityscapes $\to$ Dark Zurich to predict the test set of BDD100k-night. BDD100k-night contains 87 images with a resolution of 1280$\times$720.
		\item Nighttime Driving \cite{dai2018dark}. The Nighttime Driving includes 50 nighttime driving-scene images with a resolution of 1920$\times$1080. We also adopt it to verify the generalization of our model trained from Cityscapes $\to$ Dark Zurich.
	\end{enumerate}

	We employ per-class intersection-over-union(IoU) and mean IoU over all classes as the evaluation metric. We perform the evaluation on 19 categories for GTAV $\to$ Cityscapes, Cityscapes $\to$ Dark Zurich tasks, and both 16 and 13 categories for SYNTHIA $\to$ Cityscapes task.

	\subsection{Implementation Details}
	\subsubsection{Network architecture}
	We adopt DeepLab-V3$+$ with ResNet101 as our base segmentation architecture. Meanwhile, both the mean and covariance of the embeddings are generated by the projection head. Specifically, they are produced separately by two fully-connected layers with BN and ReLU, mapping the high-dimensional pixel embedding from the backbone into a 512-$d$ $l_2$-normalized vector. However, we replace the last ReLU function with the exponential (Exp) function when generating covariance to ensure the covariance is non-negative.  Similar to \cite{xie2023sepico}, the backbone is initialized using the weights pre-trained on ImageNet, and the other layers are initialized randomly.
	
				\begin{table}[t]
		\caption{  Hyperparameter values for the GTAV $\to$ Cityscapes (G $\to$ C), SYNTHIA $\to$ Cityscapes (S$\to$ C) and Cityscapes $\to$ Dark Zurich (C $\to$ D) tasks.}
		\label{tab1}
		\centering
		\begin{tabular}{c|ccc}
			\toprule
			Settings & G$\to$C & S$\to$C & C$\to$D \\
			\hline
			Top-$k$ ambiguity class $k$ & 10 & 10 & 10 \\
			Cropping times $N$ & 10 & 10 & 10 \\
			Total training iterations & 40k & 40k & 60k \\
			Learning rate & $6\mathrm{e}{-5}$ & $6\mathrm{e}{-5}$ & $6\mathrm{e}{-5}$\\
			$\lambda _{t}$ & 1 & 1 & 1\\
			$\lambda _{c}$ & 1 & 1 & 1\\
			$\lambda _{KL}$ &  $1\mathrm{e}{-6}$ &  $1\mathrm{e}{-6}$ &  $1\mathrm{e}{-6}$ \\
			
			\bottomrule
		\end{tabular}%
	\end{table}
	\subsubsection{Training}
	We train the network with a batch of two 640 $\times$ 640 random crops. Threshold $\alpha$ is set to 0.968. Cropping times $N$ is set to 10, the default value of $k$ for the top-$k$ ambiguity class selection is set to 10. We adopt AdamW as our optimizer with betas (0.9, 0.999) and weight decay 0.01. The initial learning rate for the backbone and projection head is set to $6\times10^{-5}$, $6\times10^{-4}$. Moreover, linear learning rate warmup and polynomial learning rate decay are used. For all experiments, we set $\lambda_{t}$, $\lambda_{c}$  to 1.0 , $\lambda_{KL}$ to $10^{-6}$, and momentum for prototype update to 0.999 respectively. We train the model for a total of 40k iterations for synthetic-to-real tasks and 60k iterations for day-to-night adaptation. The distribution contrast starts from 3k iterations. The hyperparameters for all tasks are shown in Table \ref{tab1}. We keep all hyperparameters the same as our baseline \cite{xie2023sepico}, except for $k$, which we select through experimentation. All experiments are implemented on a single NVIDIA A30 GPU with PyTorch.
	\subsubsection{Testing}
	During test time, all projection heads, the student model, and prototypes are discarded. We resize the test images to 1280$\times$640 and obtain the segmentation results directly through the adapted teacher model.


	\subsection{Comparisons with the state-of-the-arts}
	
	We compare our PPPC with the recently leading approaches in two challenging synthetic-to-real cross-domain semantic segmentation tasks including GTAV $\to$ Cityscapes and SYNTHIA $\to$ Cityscapes. Additionally, a day-time to night-time adaptation task Cityscapes $\to$ Dark Zurich is also included.
	
	Table \ref{tab2} shows the adaptation results on the task of GTAV $\to$ Cityscapes with comparisons to the state-of-the-art methods \cite{zhang2021prototypical,jiang2022prototypical,li2022class,cheng2023adpl,xie2023sepico,ren2023prototypical,zhao2023learning,li2023contrast,truong2023fredom,shen2023diga,hoyer2022hrda, fan2024otclda}. It can be seen that our PPPC achieves superior performance. Specifically, our methods achieve the best mIoU of 63.7\%, significantly outperforming the deterministic feature alignment approach SePiCo \cite{xie2023sepico}     and also surpassing OTCLDA \cite{fan2024otclda}, which aligns distributions through optimal transport. Due to the proposed probabilistic pixel embedding inherent with the capability to gather information about the style gap between two domains, thus outperforming the style transfer based method CONFETI \cite{li2023contrast} by +1.5\% and DiGA \cite{shen2023diga} which additionally combines a distillation procedure by +1.0\%. 
	Moreover, our PPPC achieves mIoU even higher than HRDA \cite{hoyer2022hrda} which uses high-resolution images as input. This shows that the proposed probabilistic pixel embedding and prototypical 
	contrast approach can effectively deal with ambiguous regions in images by considering the uncertainty of each pixel and prototype without the need for high-resolution images.
	
		\begin{table*}[!t]
		\caption{Comparison results of \textbf{GTAV} $\to$ \textbf{Cityscapes}. All methods are based on ResNet-101 for a fair comparison. The best result is highlighted in \textbf{bold},with the second best results \underline{underlined}.$^{\dagger}$  indicate method trained at higher resolution.}
		\label{tab2}
		\resizebox{\textwidth}{!}{%
			\renewcommand\arraystretch{1.5}
			\begin{tabular}{l|ccccccccccccccccccc|c}
				\toprule
				Method          & Road & S.walk & Build & Wall  & Fence  & Pole & Tr.Light & Sign & Veget. & Terrain & Sky  & Person & Rider & Car  & Truck & Bus & Train & M.bike & Bike & mIoU \\          \hline
				ProCA \cite{jiang2022prototypical} & 91.9 & 48.4 & 87.3 & 41.5 & 31.8 & 41.9 & 47.9 & 36.7 & 86.5 & 42.3 & 84.7 & 68.4 & 43.1 & 88.1 & 39.6 & 48.8 & 40.6 & 43.6 & 56.9 & 56.3 \\
				ProDA \cite{zhang2021prototypical}  & 87.8 & 56.0 & 79.7 & 46.3 & 44.8 & 45.6 & 53.5 & 53.5 & 88.6 & 45.2 & 82.1 & 70.7 & 39.2 & 88.8 & 45.5 & 59.4 & 1.0 & 48.9 & 56.4 & 57.5 \\
				ADPL \cite{cheng2023adpl}  & 93.4 & 60.6 & 87.5 & 45.3 & 32.6 & 37.3 & 43.3 & 55.5 & 87.2 & 44.8 & 88.0 & 64.5 & 34.2 & 88.3 & 52.6 & 61.8 & 49.8 & 41.8 & 59.4 & 59.4 \\
				PBAL \cite{ren2023prototypical} & 93.4 & 59.5 & 87.7 & \underline{49.7} & 41.3 & 45.2 & 53.8 & 43.7 & 88.6 & 41.7 & \textbf{91.2} & 70.5 & 36.0 & 89.4 & 49.9 & 52.1 & \textbf{57.8} & 49.0 & 46.7 & 60.4 \\
				CPSL \cite{li2022class} & 92.3 & 59.9 & 84.9 & 45.7 & 29.7 & \textbf{52.8} & \textbf{61.5} & 59.5 & 87.9 & 41.5 & 85.0 & 73.0 & 35.5 & 90.4 & 48.7 & \textbf{73.9} & 26.3 & \textbf{53.8} & 53.9 & 60.8 \\
				SePiCo \cite{xie2023sepico} & 95.2 & \underline{67.8} & 88.7 & 41.4 & 38.4 & 43.4 & 55.5 & 63.2 & 88.6 & 46.4 & 88.3 & 73.1 & 49.0 & \underline{91.4} & 63.2 & 60.4 & 0.0 & 45.2 & 60.0 & 61.0 \\
				FREDOM \cite{truong2023fredom} & 90.9 & 54.1 & 87.8 & 44.1 & 32.6 & 45.2 & 51.4 & 57.1 & 88.6 & 42.6 & 89.5 & 68.8 & 40.0 & 89.7 & 58.4 & 62.6 & \underline{55.3} & 47.7 & 58.1 & 61.3 \\
				RTea \cite{zhao2023learning} & 95.4 & 67.1 & 87.9 & 46.1 & \textbf{44.0} & 46.0 & 53.8 & 59.5 & \textbf{89.7} & \textbf{49.8} & 89.8 & 71.5 & 40.5 & 90.8 & 55.0 & 57.9 & 22.1 & 47.7 & 62.5 & 61.9 \\
				CONFETI \cite{li2023contrast} & \underline{95.7} & \underline{69.9} & \underline{89.5} & 34.6 & \underline{42.6} & 40.9 & 57.5 & 59.4 & 88.6 & \underline{49.0} & 88.2 & 72.8 & \textbf{53.4} & 90.1 & 61.8 & 54.9 & 13.9 & 50.2 & 63.4 & 62.2 \\
				DiGA \cite{shen2023diga} & 95.6 & 67.4 & \textbf{89.8} & \textbf{51.6} & 38.1 & \underline{52.0} & \underline{59.0} & 51.5 & 86.4 & 34.5 & 87.7 & \textbf{75.6} & 48.8 & \textbf{92.5} & \textbf{66.5} & \underline{63.8} & 19.7 & 49.6 & 61.6 & 62.7 \\ 
				OTCLDA \cite{fan2024otclda} & \textbf{95.8} & \textbf{70.5} & 89.1 & 40.4 & 40.6 & 45.7 & 55.7 & \textbf{67.1} & \underline{89.4} & 48.1 & \underline{90.6} &\underline{73.8} & \underline{49.3} & \underline{91.4} & \underline{66.4} & 62.1 & 0.1 & \underline{53.4} & \underline{63.5} & \underline{62.8} \\
				\hline
				PPPC(Ours)   & 94.9 & 65.2 & 88.4 & 45.5 & 41.1 & 50.1 & 56.3 & \underline{65.5} & 88.7 & 42.0 & 89.0 & 73.4 & 43.3 & 90.8 & 60.2 & 62.8 & 39.9 & 47.6 & \textbf{64.9} & \textbf{63.7} \\ \hline
				HRDA$^{\dagger}$ \cite{hoyer2022hrda} & 96.2 & 73.1 & 89.7 & 43.2 & 39.9 & 47.5 & 60.0 & 60.0 & 89.9 & 47.1 & 90.2 & 75.9 & 49.0 & 91.8 & 61.9 & 59.3 & 10.2 & 47.0 & 65.3 & 63.0  \\ \bottomrule
			\end{tabular}%
		}
	\end{table*}
	
	\begin{table*}[!t]
		\caption{Comparison results of \textbf{SYNTHIA} $\to$ \textbf{Cityscapes}. $\mathrm{mIoU^*}$ denotes the mean IoU of 13 classes excluding the classes with $^{*}$.All methods are based on ResNet-101 for a fair comparison. The best result is highlighted in \textbf{bold},with the second best results \underline{underlined}. }
		\label{tab3}
		\resizebox{\textwidth}{!}{%
			\renewcommand\arraystretch{1.5}
			\begin{tabular}{l|cccccccccccccccc|cc}
				\toprule
				Method      & Road & S.walk & Build & Wall$^\ast$ & Fence$^\ast $ & Pole$^\ast $& Tr.Light & Sign & Veget. & Sky  & Person & Rider & Car  & Bus  & M.bike & Bike & mIoU & mIoU$^\ast $\\ \hline
				DACS \cite{tranheden2021dacs}& 80.6 & 25.1 & 81.9 & 21.5 & 2.9 & 37.2 & 22.7 & 24.0 & 83.7 & \underline{90.8} & 67.6 & 38.3 & 83.0 & 38.9 & 28.5 & 47.6 & 48.4 & 54.8 \\ 
				ProCA \cite{jiang2022prototypical} & \underline{90.5} & \underline{52.1} & 84.6 & 29.2 & 3.3 & 40.3 & 37.4 & 27.3 & 86.4 & 85.9 & 69.8 & 28.7 & 88.7 & 53.7 & 14.8 & 54.8 & 53.0 & 59.6 \\
				ProDA \cite{zhang2021prototypical} & 87.8 & 45.7 & 84.6 & \textbf{37.1} & 0.6 & 44.0 & 54.6 & 37.0 & \underline{88.1} & 84.4 & 74.2 & 24.3 & 88.2 & 51.1 & 40.5 & 45.6 & 55.5 & 62.0 \\
				ADPL \cite{cheng2023adpl} & 86.1 & 38.6 &  \underline{85.9} & 29.7 & 1.3 & 36.6 & 41.3 & 47.2 & 85.0 &  90.4 & 67.5 & 44.3 & 87.4 & 57.1 & 43.9 & 51.4 & 55.9 & 63.6 \\
				DecoupleNet \cite{lai2022decouplenet}& 77.8 	&48.6& 	75.6& 	32.0 &	1.9 &	44.4 &	52.9 &	38.5 &	87.8 &	88.1 &	71.1 &	34.3 &	88.7 &	58.8 &	50.2 &	61.4 &	57.0 &	64.1 \\
				PBAL \cite{ren2023prototypical} & 86.5 & 46.5 & 85.1 & 23.6 & 0.4 & 44.1 & 49.7 & 46.5 & \textbf{88.4} & \textbf{91.6} & 72.6 & 41.5 & \underline{89.5} & \underline{63.1} & 43.1 & 48.1 & 57.5 & 65.6 \\
				SePiCo \cite{xie2023sepico} & 77.0 & 35.3 & 85.1 & 23.9 & \underline{3.4} & 38.0 & 51.0 &  55.1 & 85.6 & 80.5 & 73.5 &  \underline{46.3} & 87.6 & \textbf{69.7} & \underline{50.9} & \textbf{66.5} & 58.1 &  \underline{66.5} \\
				CPSL \cite{li2022class} & 87.2 & 43.9 &  85.5 & \underline{33.6} & 0.3 & \underline{47.7} & \textbf{57.4} & 37.2 & 87.8 & 88.5 & \textbf{79.0} & 32.0 & \textbf{90.6} & 49.4 &  50.8 & 59.8 &  58.2 & 65.3 \\
				OTCLDA \cite{fan2024otclda} & 87.3 & 45.9 & \textbf{87.0}	& 17.0 & \textbf{5.7} & \textbf{49.0} & \underline{55.9} & \underline{56.7} & 81.6 & 75.7 & \underline{74.6} & 41.4 & 89.0 & 53.7 & \textbf{51.1} & \underline{63.8} & \underline{58.5} & 66.4 \\
				\hline
				Ours &   \textbf{90.9} & \textbf{52.9} & 84.7 & 21.9 & 2.1 & 45.4& 49.4 & \textbf{57.0} & 85.2 & 87.5 & 72.9 &  \textbf{46.3} & 86.8 & 47.2 & 50.4 &  61.0 & \textbf{58.8} & \textbf{67.1}
				\\ \bottomrule     
			\end{tabular}%
		}
	\end{table*}
	
	\begin{table*}[!t]
		\caption{Comparison results of \textbf{Cityscapes} $\to$ \textbf{Dark Zurich}. All methods are based on ResNet-101 for a fair comparison. The best result is highlighted in \textbf{bold},with the second best results \underline{underlined}.}
		\label{tab4}
		\resizebox{\textwidth}{!}{%
			\renewcommand\arraystretch{1.5}
			\begin{tabular}{l|ccccccccccccccccccc|c}
				\toprule
				Method & Road & S.walk & Build & Wall & Fence & Pole & Tr.Light & Sign & Veget. & Terrain & Sky & Person & Rider & Car & Truck & Bus & Train & M.bike & Bike & mIoU \\ \hline
				DANNet \cite{wu2021dannet} & 90.4 & 60.1 & 71.0 & 33.6 & 22.9 & 30.6 & 34.3 & 33.7 & \underline{70.5} & 31.8 & 80.2 & 45.7 & 41.6 & 67.4 & 16.8 & 0.0 & 73.0 & 31.6 & 22.9 & 45.2 \\
				DIAL-Filters \cite{liu2023improving} & 90.6 & 60.8 & 70.9 & \underline{40.2} & 21.1 & 39.6 & 34.4 & \underline{38.3} & \textbf{73.2} & 30.2 & 72.9 & \textbf{48.6} & 41.6 & \underline{72.8} & 8.8 & 0.0 & \textbf{74.6} & 33.0 & 22.8 & 46.0 \\
				LoopDA \cite{shen2023loopda} & 86.3 & 46.3 & \textbf{76.1} & 30.3 & 22.5 & 32.5 & 34.1 & 34.8 & 62.6 & 19.5 & \underline{84.3} & \underline{46.6} & 51.5 & \textbf{73.2} & \underline{60.7} & \underline{3.1} & \underline{73.4} & 26.2 & 24.8 & 46.8 \\
				DANIA \cite{wu2021one} & 90.8 & 59.7 & 73.7 & 39.9 & \underline{26.3} & 36.7 & 33.8 & 32.4 & \underline{70.5} & \underline{32.1} & \textbf{85.1} & 43.0 & 42.2 & \underline{72.8} & 13.4 & 0.0 & 71.6 & \textbf{48.9} & 23.9 & 47.2 \\
				CCDistill \cite{gao2022cross} & 89.6 & 58.1 & 70.6 & 36.6 & 22.5 & 33.0 & 27.0 & 30.5 & 68.3 & \textbf{33.0} & 80.9 & 42.3 & 40.1 & 69.4 & 58.1 & 0.1 & 72.6 & \underline{47.7} & 21.3 & \underline{47.5} \\
				\hline
				SePiCo \cite{xie2023sepico} & \textbf{91.2} & \underline{61.3} & 67.0 & 28.5 & 15.5 & \underline{44.7} & \textbf{44.3} & \textbf{41.3} & 65.4 & 22.5 & 80.4 & 41.3 & \textbf{52.4} & 71.2 & 39.3 & 0.0 & 39.6 & 27.5 & \underline{28.8} & 45.4 \\ 
				PPPC(Ours)  & \underline{90.9} & \textbf{63.0} & \underline{75.6} & \textbf{43.9} & \textbf{27.6} & \textbf{53.3} & \underline{ 40.9} & 33.1 & 67.1 & 29.2 & 79.8 & 46.3 & \underline{51.7} & 53.6 & \textbf{61.6} & \textbf{14.2} & 65.0 & 33.6 & \textbf{30.8} & \textbf{50.6} 
				\\ \bottomrule     
			\end{tabular}%
		}
	\end{table*}
	
	Table \ref{tab3} shows the adaptation results on the task SYNTHIA $\to$ Cityscapes. This adaptation task is more challenging than the previous one due to the large domain gap, but our PPPC still achieves significant improvements over competing methods. PPPC attains 58.8\% mIoU and 67.1\% mIoU$^\ast$ achieving a significant improvement with ProDA \cite{zhang2021prototypical} and PBAL \cite{ren2023prototypical}, outperforming them by +3.2\% mIoU and +1.2\% mIoU respectively. Our method can outperform these approaches without the need for any complex strategies like muti-stage self-training in ProDA or knowledge distillation in PBAL, which further verifies the benefits of distributional pixel contrast with prototypes. Notably, our methods achieve the best mIoU$^\ast$ at 67.1\%, surpassing the approach CPSL \cite{li2022class}, which addresses class imbalance and refines pseudo-labels through clustering pixels, further highlighting the superiority of distributional embedding under class imbalance scenarios. Additionally, the higher performance compared to OTCLDA \cite{fan2024otclda} demonstrates that our kernel function is more advantageous than optimal transport (i.e., Wasserstein Distance). A detailed comparison can be found in Section 4.4.5.

	Table \ref{tab4} shows the adaptation results on the task Cityscapes $\to$ Dark Zurich. Our PPPC achieves significant improvement over the methods explicitly tailored for addressing day-time to night-time adaptation, outperforming CCDstill \cite{gao2022cross} by +3.1\% mIoU and LoopDA \cite{shen2023loopda} by +3.8\%. Like SePiCo \cite{xie2023sepico}, our PPPC is a universal framework for domain adaptation, not limited to addressing day-time to night-time tasks. Compared to SePiCo, we not only exhibit noticeable improvement in majority groups such as buildings, vegetation, and person but also achieve significant enhancement in minority groups such as fences, poles, and trucks, eventually leading gain of +5.2\%. 
	
	\begin{figure*}[!t]
		\centering
		\includegraphics[width= \textwidth]{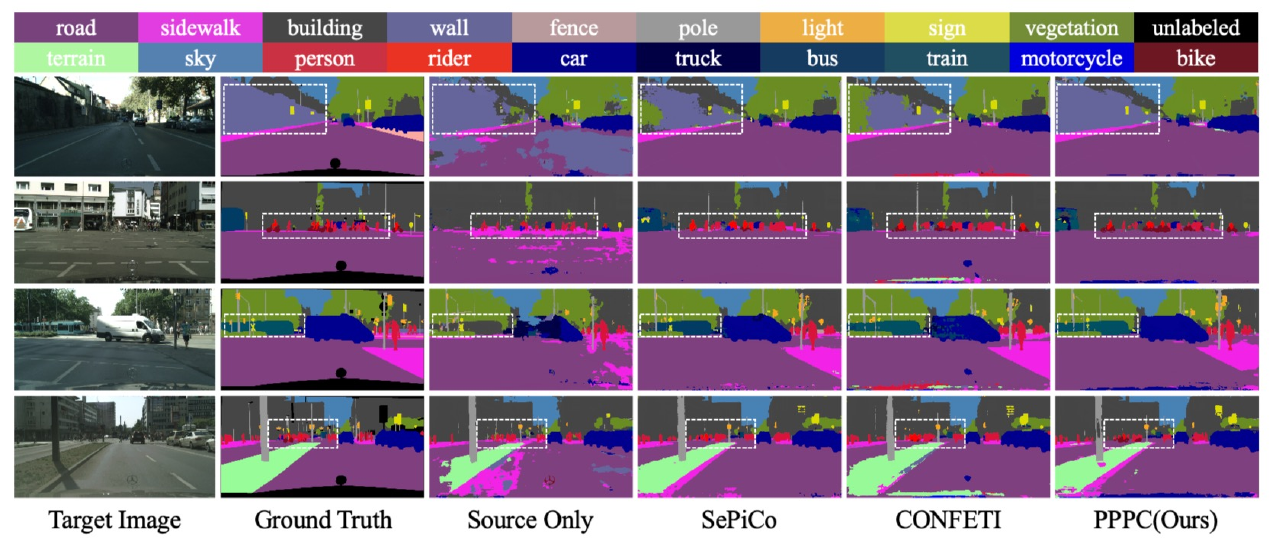}
		\caption{Qualitative results of domain adaptive semantic segmentation task GTAV $\to$ Cityscapes. Better segmentation results are highlighted in dash boxes. PPPC improves the segmentation of ambiguous scenarios such as inadequately illuminated walls, crowded persons, and challenging classes like bus.}
		\label{fig5}
	\end{figure*}
	
	Fig. \ref{fig5} shows the qualitative segmentation results on Cityscapes. We compare our results predicted by PPPC to the others obtained by the Source Only, SePiCo \cite{xie2023sepico}, and CONFETI \cite{li2023contrast}. Our probabilistic methods PPPC predict a clearer outline in ambiguous objects like backlit walls, dense pedestrian crowds, overlapping bicycles, and distant traffic signs, showing a significant improvement over the deterministic methods.
	
	\subsection{Ablation study} 
	
	\begin{table}[]
		\caption{Ablation study on the GTAV $\to$ Cityscapes (G $\to$ C) and  SYNTHIA $\to$ Cityscapes tasks(S$\to$ C). AGC denotes Ambiguity-guided Cropping.  }
		\label{tab5}
		\centering
		\begin{tabular}{c|cccc|cc|cc}
			\toprule
			\multirow{2}{*}{Method}	& \multirow{2}{*}{$\mathcal{L} _{ssl}$} & \multirow{2}{*}{$\mathcal{L} _{cl}$} & \multirow{2}{*}{$\mathcal{L} _{KL}$} & \multirow{2}{*}{AGC} &  \multicolumn{2}{c}{G $\to$ C} &   \multicolumn{2}{c}{S$\to$ C} \\ \cline{6-9}
			& 	&	&	&	&	mIoU & $\Delta$  &	mIoU & $\Delta$\\
			\hline
			\multirow{4}{*}{w/o self-training} &  &  &  &  & 37.1 & - & 26.5&  -\\
			&  & \cmark  &  &  & 43.5 & 6.4  & 29.2& 2.7\\ 
			&  & \cmark  & \cmark  &  & 45.2 & 8.1   & 30.3&3.8\\
			&  & \cmark  & \cmark  & \cmark  & 47.2 & 10.1   & 32.5&6.0\\
			\hline
			\multirow{4}{*}{PPPC(Ours)} & \cmark &  &  &  & 59.4 & -  & 55.1& -\\
			& \cmark  & \cmark  &  &  & 58.7 & -0.7   & 53.3&-1.8\\
			& \cmark  & \cmark  & \cmark  &  & 62.0 & 2.6  & 56.1&1.0\\
			& \cmark  & \cmark  & \cmark  & \cmark  & \textbf{63.7} & 4.3 & \textbf{58.9}&3.8\\
			\bottomrule
		\end{tabular}%
	\end{table}
	
	We conduct comprehensive ablation studies on each component present in the proposed method. For a fair comparison, all experiments are implemented under the same training settings.
	\begin{enumerate}[wide]
		\item \textit{\textbf{Effect of probabilistic prototypical pixel contrast}}. As mentioned before, the probabilistic contrast not only reduces the intra-class distance and increases inter-class distance, which is easy to realize by deterministic contrast, but makes the decision boundary clearer near the ambiguous embedding point. We investigate the effect of probabilistic contrast with and without the self-training framework separately. As shown in Table \ref{tab5}, probabilistic contrast brings mIoU gains of +6.4\% on  GTAV $\to$ Cityscapes and +2.7\% on SYNTHIA $\to$ Cityscapes without self-training which is less than using self-training techniques alone. Interestingly, when we combine the probabilistic contrast with self-training, the performance is slightly degraded compared with the result using only self-training. We conjecture that the model trained without self-training lacks the ability to correlate the two domains, hence resulting in a substantial mean difference ,i.e., $(\mu_1-\mu_2 )^2$ , which can be easily addressed by probabilistic contrast. However, the self-training strategy boosts the transfer capability of the model between two domains, leading to relatively lower values on the numerator in Eq. \eqref{eq15}. which could cause erroneous optimization direction if the covariance on the denominator and logarithm is not properly regulated.

		\item \textit{\textbf{Effect of $\mathcal{L}_{KL}$}}. Because our loss function is left-unbounded, an incorrect optimization direction results in the covariance converging to $\infty$. We study the advantages of using KL divergence as a regularization term on the similarity measurement of two distributions in Table \ref{tab5}. It is obviously observed that by adding the KL divergence, the covariance is limited to a meaningful range, and the full potential of probabilistic contrast is released, achieving a significant gain of +1.7\% and +3.3\% on  GTAV $\to$ Cityscapes, +1.1\% and +2.8\% on SYNTHIA $\to$ Cityscapes respectively. 
		
		\begin{figure}[t]
			\centering
			\includegraphics[width=0.6 \linewidth]{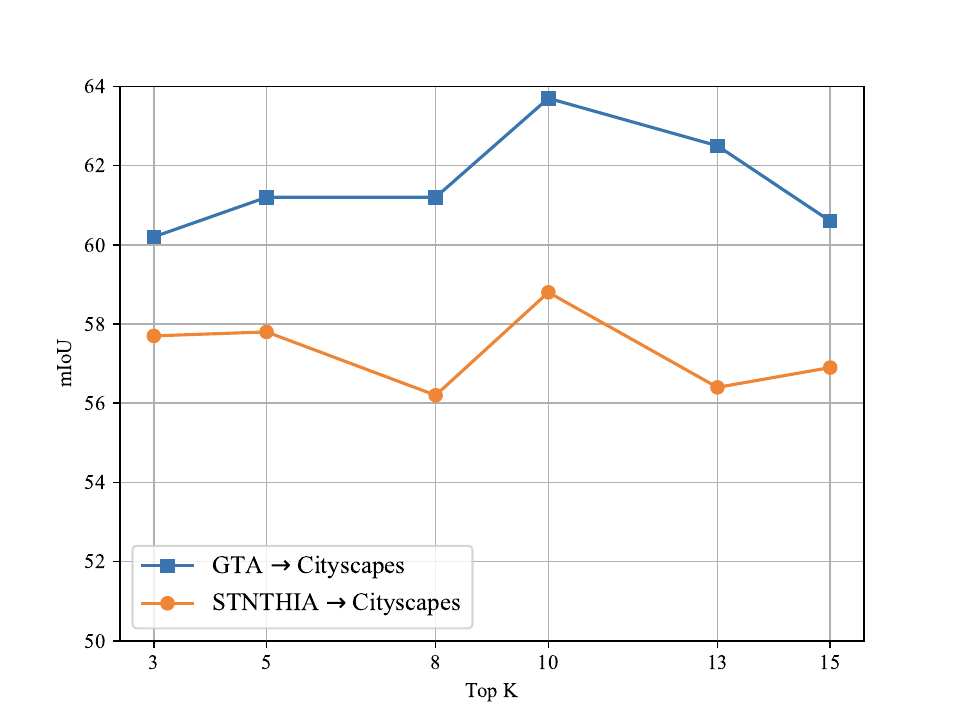}
			\caption{Parameter sensitivity analysis for AGC.}
			\label{fig6}
		\end{figure}
		
		\begin{table}[]
			\caption{Comparison results of mIoU for different cropping methods on the target domain.  }
			\label{tab6}
			\centering
			\begin{tabular}{c|cccc}
				\toprule
				Method & Random & CBC & AGC (Constant) & AGC   \\
				\hline
				
				mIoU & 61.4  & 60.5  & 61.6    & \textbf{63.7}  \\
				\bottomrule
			\end{tabular}%
		\end{table}
		
		\item \textit{\textbf{Effect of ambiguity-guided cropping(AGC)}} . As shown in Table \ref{tab5}, our cropping method achieves noticeable gains of +2.0\% and +1.7\% on  GTAV $\to$ Cityscapes, +2.2\% and +2.8\% on SYNTHIA $\to$ Cityscapes. The results imply that the covariance predicted by the model not only benefits the alignment at the feature level but also helps to select reasonable inputs according to the transfer capability of the model dynamically at the data level. Next, we analyze the parameter sensitivity of AGC, and compare it with other strategies on target domains. From Fig. \ref{fig6}, we can see that increasing the value of $k$ within the range of not greater than 10 slightly improves the performance, but the performance drops when $k$ is greater than 10. We conjecture that although the precise trend is not aligned due to different source domains (e.g. the STNTHIA has 16 classes which is fewer than 19 classes in GTAV), the head classes with the highest covariance are generally the same. Thus, we choose 10 as the default value of $k$ for all experiments.
		
		Furthermore, Table \ref{tab6} shows the performance of different cropping methods implemented on target domain inputs. Random stands for cropping a image to given size randomly, CBC  stands for class-balanced cropping which gives crops by evaluating the max ratio of each class and the score of classes meet the condition, AGC (Constant) stands for the same configuration with AGC except we fixed the covariances of prototypes to the reciprocal of the class frequency in the source dataset. As seen, CBC achieves the worst result even lower than random selection, showing that CBC is not suitable for probabilistic-based methods. Moreover, AGC (Constant) exhibits a marginal performance gap + 0.2\% with random  selection, indicating that our probabilistic embedding has, to some extent, addressed the class imbalance issue, the model will no longer benefit from continuing to use class-balance strategies. Thus, the main issue with these methods lies in their inability to effectively introduce ambiguous points near the decision boundary at the data level. Conversely, our AGC is designed to tackle this problem by scoring each crop with class-wise uncertainty provided by prototypes. With dynamically adjusted crop output by our AGC strategy, we achieve +2.3\% mIoU compared with random selection.

		\begin{figure}[t]
			\centering
			\includegraphics[width=\linewidth]{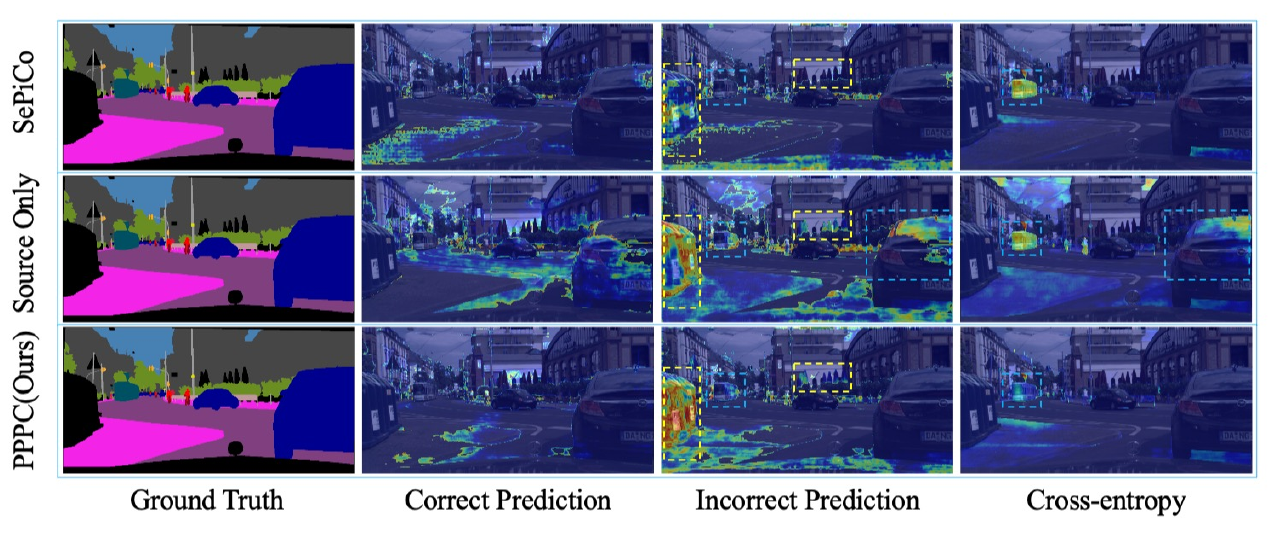}
			\caption{Visualization of the entropy map for correct predictions, incorrect predictions, and the cross-entropy map with ground truth. PPPC shows low uncertainty for correct predictions, and high uncertainty for both unlabeled and incorrect classes. For incorrect predictions, apart from the unlabeled classes, the more consistent it is with corss-entropy map, the model is less likely to assign high confidence to the incorrect class.}
			\label{fig7}
		\end{figure}
		
		\begin{table}[]
			\caption{The degree of freedom (DoF) for $\Sigma$ and mIoU on GTAV $\to$ Cityscapes (G $\to$ C) and  SYNTHIA $\to$ Cityscapes tasks(S$\to$ C).}
			\label{tab7}
			\centering
			\begin{tabular}{c|ccc}
				\toprule
				Method & DoF($\sigma$) & G $\to$ C & S $\to$ C    \\
				\hline
				
				PPPC $\mu$ only & 0  & 59.4  & 56.0     \\
				PPPC iostropic & 1  & 59.1  & 56.6     \\
				PPPC  & 512  & \textbf{63.7}  & \textbf{58.8}     \\
				\bottomrule
			\end{tabular}%
		\end{table}
		
		\item \textit{\textbf{Effect of probabilistic embedding}}. We model the probabilistic embedding with both $\mu$ and $\Sigma$. We remove the covariance $\Sigma$ to illustrate its necessity. As shown in Table \ref{tab7}, DoF denotes the degree of freedom for covariance. Removing the covariance of embedding leads our method to degrade to a deterministic approach, resulting in a decrease of -4.3\% and -2.8\% on GTAV$\to$ Cityscapes and SYNTHIA $\to$ Cityscapes, demonstrating the significance of modeling covariance in probabilistic methods. Furthermore, although we parameterize the covariance matrix $\Sigma\in\mathbb{R}^{D\times D}$ with its diagonal element vector $\sigma\in\mathbb{R}^D$, one may consider modeling the covariance $\Sigma$ with the same value on each position of diagonal elements, leading to isotropic gaussian distributions. From Table \ref{tab7}, we can observe that increasing the freedom of covariance $\Sigma$ brings continuous improvement, however, to increase the training efficiency, we just choose 512-$d$ for all experiments.
		
		To illustrate how our model leverages uncertainty, we visualize the entropy map for both correct and incorrect predictions, as well as the cross-entropy map in Fig. \ref{fig7}. Entropy map for predictions is derived from the calculation of max probability predicted from the unlabeled image for each pixel, describing the model’s uncertainty for each part of the image. On the other hand, the cross-entropy map is generated based on the cross-entropy between the model’s prediction and the ground truth label, depicting the extent of deviation from the correct results. For a discriminative model, lower uncertainty should be assigned to correct predictions, while higher uncertainty for incorrect predictions. From these visualizations, we can observe that (i) for correct predictions, our PPPC shows less uncertainty about each pixel; (ii) for incorrect predictions, by comparing with the cross-entropy map, the deterministic approach SePiCo is prone to assign high confidence to the wrong class (marked as the blue box), which would lead to incorrect optimization direction as shown in Fig. \ref{fig1}(b). However, our probabilistic method, even the source-only version, is capable of giving high uncertainty to ambiguous pixels, ensuring most of the incorrect predictions are among the high-uncertainty groups; (iii) Although the unlabeled class is not taken into account in the cross-entropy calculation, our method still can discriminative it from the other class, and assign high uncertainty to it (marked as the yellow box), which is struggled for deterministic method. This further indicates the generalization of our PPPC and its potential application in risk control for autonomous driving.

		\begin{table}[]
			\caption{Comparison results of mIoU for different distribution measurements trained on GTAV $\to$ Cityscapes.The best result is highlighted in \textbf{bold},with the second best results \underline{underlined}, the third best results \underline{\underline{double underlined}}.  }
			\label{tab8}
			\centering
			\begin{tabular}{c|cc}
				\toprule
				Method  & Sampling & mIoU \\ \hline
				\multirow{5}{*}{Cosine Similarity}  & 10 & 62.0 \\
				&   20 & 62.3 \\
				&   30 & \underline{63.4} \\
				&   40 & 61.7 \\
				&   50 & 60.8 \\ \hline
				KL Divergence & \xmark & 60.1 \\
				JS Divergence  & \xmark & 60.0 \\
				Wasserstein   Distance   & \xmark & 60.4 \\ \hline
				Bhattacharyya Kernel  & \xmark & \underline{\underline{62.7}} \\
				Expected Likelihood Kernel &   \xmark & \textbf{63.7}\\
				\bottomrule
			\end{tabular}
		\end{table}

		\item \label{distribution} \textit{\textbf{Effect of distribution measurement}}. To demonstrate the effectiveness and efficiency of our method, we compare it with other probabilistic distance variants to measure the distance between two Gaussian distributions. One may consider using the Monte-Carlo sampling strategy to approximate the similarity between two distributions as:
		\begin{equation}
			\label{eq21}
			sim(p(z_i|x_i),p(z_i|\rho_i))=\frac1{J^2}\sum_{i=1}^J\sum_{j=1}^Jd(z_i^{x_i},z_j^{\rho_j}),
		\end{equation}
		where $p(z_i|x_i)$ denotes the probability of pixel embedding, $p(z_i|\rho_i)$) denotes the embedding of prototype, $J$ is the sampling number for each distribution and $d(\cdot, \cdot)$ refers to cosine similarity. We also compare with divergence-based methods, including KL divergence, JS divergence, and Wasserstein Distance. Furthermore, we introduce another special case of PPK when $\rho=\frac{1}{2}$ , it is called Bhattacharyya Kernel (BK):
		\begin{align}
			\label{eq22}
			BK(p,q)=-\frac D2\mathrm{log}\frac12+\frac12\mathrm{log~}(\frac{\Sigma_1\Sigma_2}{\Sigma_1^{\frac{1}{2}}+\Sigma_2^{\frac{1}{2}}})-\frac14\frac{(\mu_1-\mu_2)^2}{\Sigma_1+\Sigma_2}.
		\end{align}
		As shown in Table \ref{tab8}, it can be observed that the divergence-based method performs the worst and is numerically unstable during the experiments.  This instability arises because it involves division terms by variances $\Sigma_1$ and $\Sigma_2$ (i.e., $\frac{\Sigma_1}{\Sigma_2}$), which can lead to numerical instability when the variances are very small. That is to say, when we have a highly certain embedding with nearly zero variance, the KL divergence between $p$ and $q$ will explode. Increasing the number of samples helps to improve the performance of measuring the distribution similarity. Notably, when the number of samples is set to 30, it achieves a comparable performance to kernel methods. However, it significantly increase computation overhead and the space complexity is $O(NJ^2)$, where $N$ refers the number of pixel embeddings. Instead, our kernel methods achieve the best results with space complexity of only $O(2N)$ and does not require complex operations such as sampling and reparameterization.
		
		\begin{figure}[t]
			\centering
			\includegraphics[width=\linewidth]{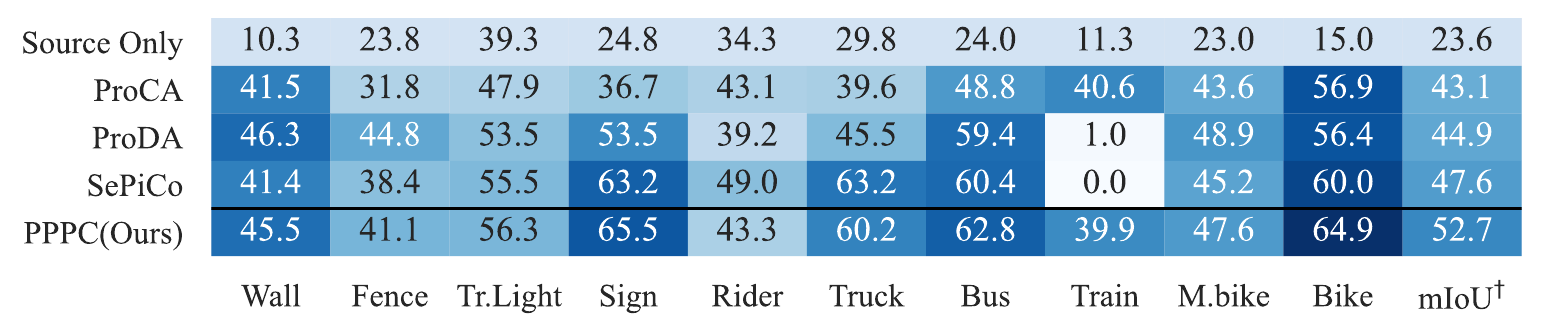}
			\caption{The results of different methods on long-tail classes. $\mathrm{mIoU}^\dagger$ indicates mIoU for tail classes.}
			\label{fig8}
		\end{figure}
		
		\item \textit{\textbf{Effect of combination of all components}}. When we combine all ingredients properly in a unified pipeline, the full potential of our model is released, resulting a 63.7\% mIoU surpassing even the generative-based method CONFETI \cite{li2023contrast}. Our method has not only achieved considerable improvement in the head class but also notable enhancements in the tail class. As shown in Fig. \ref{fig8}, the majority of improvement lies in easily confused pairs like train and bus, motorbike and bike. Additionally, smaller objects such as fences, traffic signs, and traffic lights also show a slight increase, achieving a 52.7\% tail-class $\textrm{mIoU}^{\dagger}$.
	\end{enumerate}
	\subsection{t-SNE visualization}
	
	\begin{figure}[t]
		\centering
		\includegraphics[width=\linewidth]{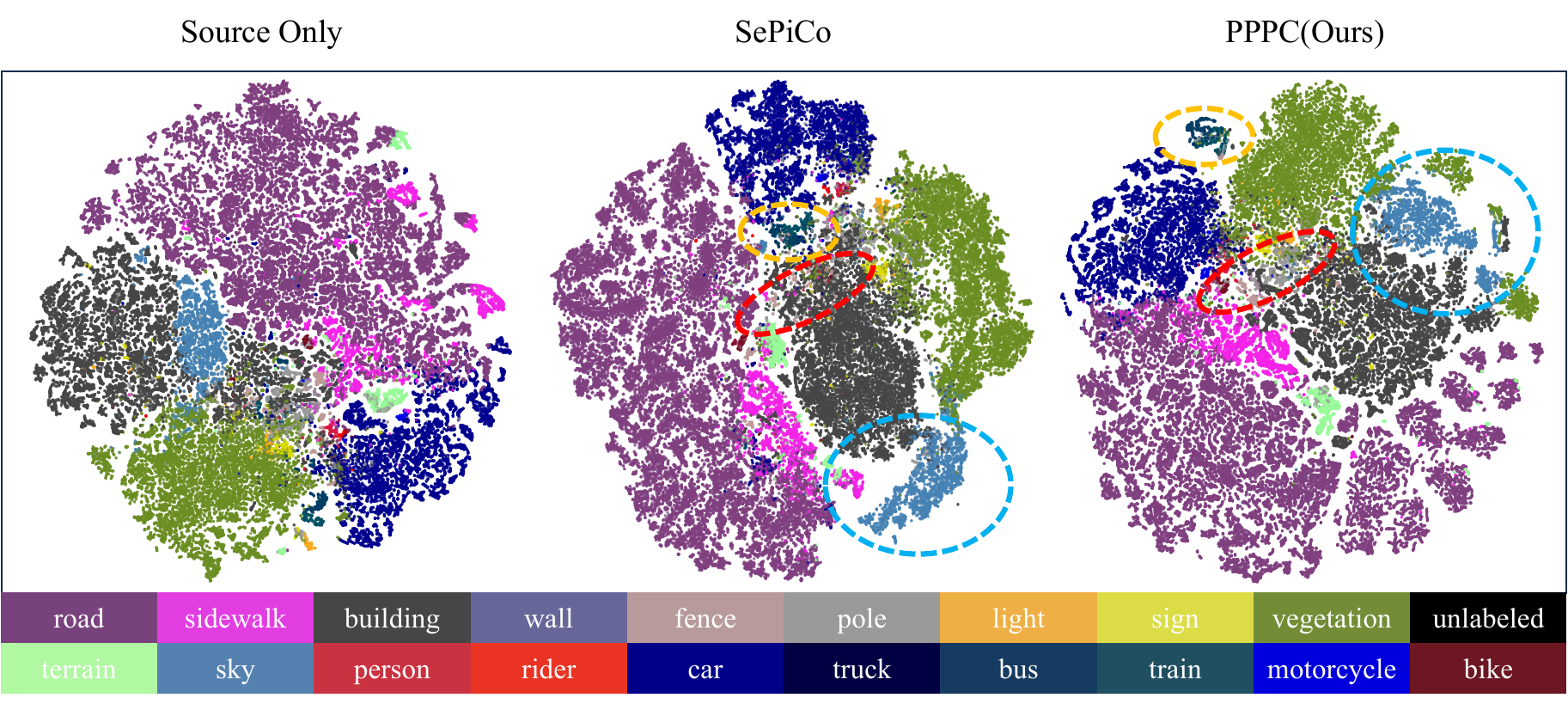}
		\caption{T-SNE analysis for the bottleneck features for ResNet-101 on the Cityscapes val. set.}
		\label{fig9}
	\end{figure}
	
	To gain insight on the improved discriminative capability, Fig. \ref{fig9} visualizes the features of the deterministic method SePiCo, our probabilistic method, and source only method for reference. We select a few images from the target domain to ensure that all 19 classes are represented. Then we map the feature representations, derived from the final stage of ResNet-101 backbone, also served as input to the decoder, into 2D space to create t-SNE visualization. We can observe that SePiCo tends to treat a class as a single cluster, employing contrastive learning to make the intra-class compact and inter-class distant. Conversely, our PPPC employs a probabilistic approach, resulting in multiple clusters within the same class, preserving the diversity of features in each class. Additionally, it is noteworthy that we successfully distinguish between train and bus, which are easily confused by other methods (marked as the orange circle). Interestingly, it seems that SePiCo has a more distinct boundary than our method (e.g., sky class exhibits a far distance from vegetation and building, marked as the blue circle), however, its mIoU performance is lower than our method. Hence, we conjecture that intra-class compactness and inter-class separability are not necessary for enhancing model discriminative capability, while handling boundary points properly might be the most crucial, which aligns with the core idea of our proposed method as shown in Fig. \ref{fig1}(c). Moreover, this assumption is further validated by our PPPC, which significantly reduces the overlapping area (marked as the red circle).
	
	\begin{table}[]
		\caption{Comparison results of Cityscape $\to$ Dark Zurich trained models for generalization on two unseen target domains: Nighttime Driving and BDD100k-night test sets. All methods are based on ResNet-101 for a fair comparison. }
		\label{tab9}
		\resizebox{\columnwidth}{!}{%
			\centering
			\begin{tabular}{c|cccc}
				\toprule
				Method & Dark Zurich & Nighttime Driving & BDD100k-night & Cityscapes \\ \hline
				DANNet  \cite{wu2021dannet}& 45.2 & 47.7 & - & - \\
				DANIA \cite{wu2021one}& - & \underline{48.4} & - & - \\
				MGCDA \cite{sakaridis2020map}& 42.5 & \textbf{49.4} & \textbf{34.9} & - \\
				CCDistill \cite{gao2022cross}& \underline{47.5} & - & \underline{33.0} & - \\ \hline
				SePiCo \cite{xie2023sepico}& 45.4 & 47.0 & 22.4 & \underline{74.4} \\
				PPPC(Ours) & \textbf{50.6} & 47.6 & 24.5 & \textbf{77.3} \\
				\bottomrule
			\end{tabular}%
		}
	\end{table}
	
	\subsection{ Generalization to unseen domains}
	We also validate the generalization of PPPC, the model trained on Cityscapes $\to$ Dark Zurich is directly tested on two unseen domains, Nighttime Driving and BDD100k-night. In Table \ref{tab9}, we can see that our PPPC achieves a comparable performance, although slightly lower than the method carefully designed for night-time adaptation tasks. Moreover, our PPPC outperforms SePiCo on the source domain, indicating that our probabilistic approach is not only beneficial to the target domain and target-related domains but also contributes to the model’s discriminative capability on the source domain.
	
	\section{Efficiency Analysis}
	
	\subsection{Relationship between embedding norms and variance}
	
	\begin{figure}[t]
		\centering
		\includegraphics[width=\linewidth]{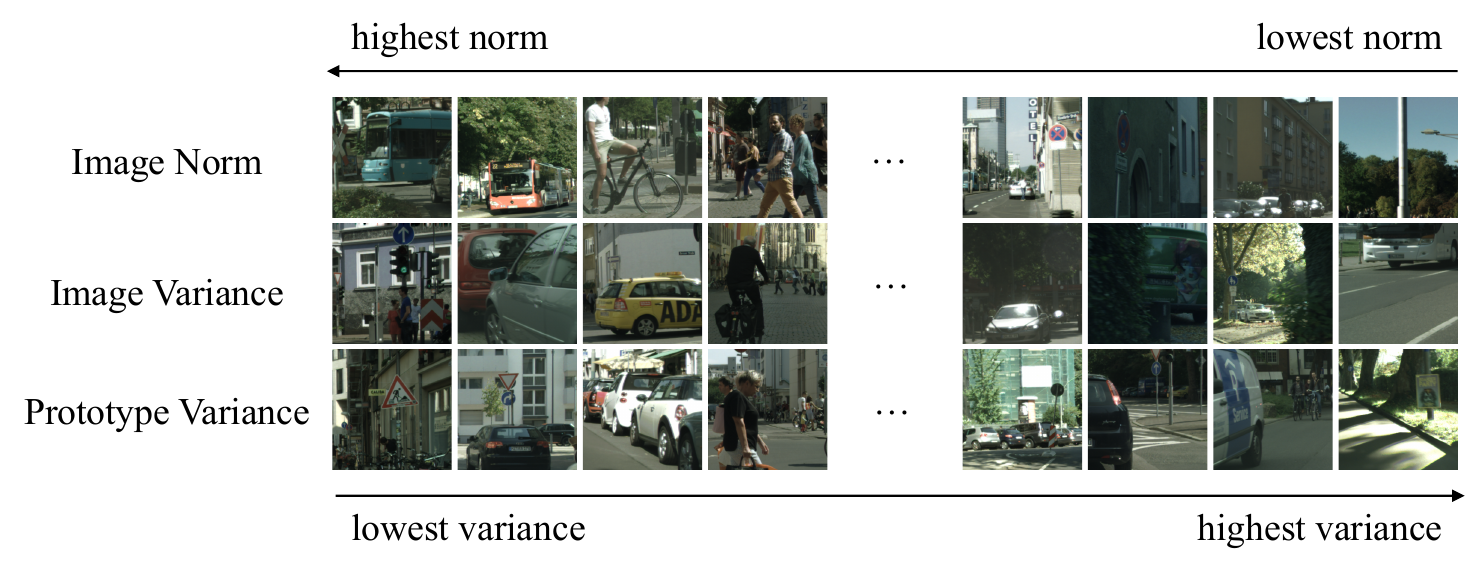}
		\caption{Cityscapes val. image crops with lowest(right) to highest(left) embedding norms and crops with lowest(left) to highest(right) variances. It can be observed that crops with high variance or low norm show the same trend, both involving ambiguous scenarios, vice versa.}
		\label{fig10}
	\end{figure}
	
	We further analyze the relationship between norms and variance to highlight the superiority of our method. We calculate image norms for each crop by averaging the $l_2$-norms of all the pixel embeddings. Fig. \ref{fig10} shows the images with the highest and lowest embedding norm in the validation set. In high-norm images, the characteristic parts of objects are particularly prominent, facilitating the detection of class-discriminative features. However, for samples with low norm, illumination appears dim and objects are overlaid, hindering the learning, in line with recent findings \cite{scott2021mises}. Similarly, we computed image variance and prototype variance respectively, observing a consistent trend between low variance and high norms images, and vice versa. Therefore, we believe that variance serves as another means to describe image ambiguity, much like the role of norms. More importantly, the computation of prototypes is independent of the network’s forward propagation (i.e.,  Eq. \eqref{eq12}), unlike image norms and variance, which require obtaining the values through the entire backbone. Thus, our AGC poses as an efficient approximation for selection via image norms. 
	
	\subsection{Throughout}
	
	\begin{table}[]
		\caption{Efficiency comparison with different methods on GTAV $\to$ Cityscapes. `Memory' represents GPU memory usage, `Time' represents the average training time per iteration, `Params' denotes trainable parameters, and `FLOPs' denotes floating point operations. Results are obtained using NVIDIA A30 with a batch size of 2. } 
		\label{tab10}
		\resizebox{\columnwidth}{!}{%
			\centering
			\begin{tabular}{c|ccccc}
				\toprule
				Method & Memory (GB) & Time (s) & Params (M)& FLOPs (G)&mIoU    \\
				\hline
				
				Source Only & 10.9   & 1.3  & 353.5& 41.2 &37.1     \\ \hline
				SePiCo \cite{xie2023sepico}& 12.2   & 3.1  &645.2& 49.1 & 61.0     \\
				HRDA \cite{hoyer2022hrda}& 22.1   & 3.2  & 734.0& 43.9 & 63.0      \\
				PPPC(Ours) & 15.2   & 3.3  & 713.1& 50.2 &\textbf{63.7}      \\
				\bottomrule
			\end{tabular}%
		}
	\end{table}
	
	The efficiency comparison of representative methods are shown in Table \ref{tab10}. We can observe that PPPC, unlike HRDA, doesn’t require the use of high-resolution images, which nearly doubles the GPU memory footprint. It only introduces a slight increase in computation and memory usage compared to SePiCo but achieves a significant enhancement. Moreover, compared to deterministic methods like SePiCo, our probabilistic approach adds only an additional probability head to encode means and variances, leading to a relative increase in the number of parameters. However, because we use posterior probability estimates for prototypes and PPK for contrastive learning, the FLOPs increases only slightly despite the increase in parameters.
	
	\section{Limitations and future work}
	Our model has demonstrated its efficacy; however, like most existing approaches, it also has some limitations. For example, we assume all observations are conditionally independent given the prototypes in Eq. \eqref{eq9}. This assumption may not be appropriate in practical scenarios. Additionally, our method only addresses inter-class confusion issues and does not improve the decoder structure, resulting in somewhat rough segmentation results for object boundaries. Finally, our method cannot yet be extended to Transformer models because the KL divergence cannot constrain the variance to a reasonable range, leading to training instability and early collapse.
	
	In future work, we will explore how to use multi-task learning (MTL) to stabilize pixel embedding modeling and extend it to a wider range of backbone architectures. Additionally, we will investigate decoder structures that facilitate fine-grained segmentation in the context of UDA.
	
	\section{Conclusion}
	In this paper, we propose PPPC, a universal probabilistic adaptation framework tailored for semantic segmentation, which successfully addresses the issue of ambiguous classes commonly existing in the self-training paradigm for UDA. Specifically, we model each pixel embedding as a multivariate Gaussian distribution, and prototypes are computed based on the observation of these embeddings, preserving both the semantic diversity for each class and the uncertainty for each embedding. Moreover, an effective and efficient metric for measuring the similarity between two distributions is adopted, which helps the model better deal with ambiguous embeddings. Additionally, a sampling strategy, AGC, which obtains class-wise uncertainty from online-updated prototypes, is employed to increase the number of ambiguous embeddings at the input level. Extensive experiments demonstrate the superiority of PPPC in both synthetic-to-real and day-to-night adaptation tasks.
	
	\section*{Acknowledgements}
	This work was supported in part by the National Natural Science Foundation of China under Grants 62276088 and 62102129.




%
%
%
\bibliographystyle{elsarticle-num}  
\bibliography{bib1}  

\begin{thebibliography}{10}
\expandafter\ifx\csname url\endcsname\relax
  \def\url#1{\texttt{#1}}\fi
\expandafter\ifx\csname urlprefix\endcsname\relax\def\urlprefix{URL }\fi
\expandafter\ifx\csname href\endcsname\relax
  \def\href#1#2{#2} \def\path#1{#1}\fi

\bibitem{janai2020computer}
J.~Janai, F.~G{\"u}ney, A.~Behl, A.~Geiger, et~al., Computer vision for
  autonomous vehicles: Problems, datasets and state of the art, Foundations and
  Trends{\textregistered} in Computer Graphics and Vision 12~(1--3) (2020)
  1--308.

\bibitem{ronneberger2015u}
O.~Ronneberger, P.~Fischer, T.~Brox, U-net: Convolutional networks for
  biomedical image segmentation, in: Medical Image Computing and
  Computer-Assisted Intervention--MICCAI 2015: 18th International Conference,
  Munich, Germany, October 5-9, 2015, Proceedings, Part III 18, Springer, 2015,
  pp. 234--241.

\bibitem{cordts2016cityscapes}
M.~Cordts, M.~Omran, S.~Ramos, T.~Rehfeld, M.~Enzweiler, R.~Benenson,
  U.~Franke, S.~Roth, B.~Schiele, The cityscapes dataset for semantic urban
  scene understanding, in: Proceedings of the IEEE/CVF Conference on Computer
  Vision and Pattern Recognition, 2016, pp. 3213--3223.

\bibitem{chen2018encoder}
L.-C. Chen, Y.~Zhu, G.~Papandreou, F.~Schroff, H.~Adam, Encoder-decoder with
  atrous separable convolution for semantic image segmentation, in: Proceedings
  of the European Conference on Computer Vision, 2018, pp. 801--818.

\bibitem{chen2017deeplab}
L.-C. Chen, G.~Papandreou, I.~Kokkinos, K.~Murphy, A.~L. Yuille, Deeplab:
  Semantic image segmentation with deep convolutional nets, atrous convolution,
  and fully connected crfs, IEEE Transactions on Pattern Analysis and Machine
  Intelligence 40~(4) (2017) 834--848.

\bibitem{hoyer2022daformer}
L.~Hoyer, D.~Dai, L.~Van~Gool, Daformer: Improving network architectures and
  training strategies for domain-adaptive semantic segmentation, in:
  Proceedings of the IEEE/CVF Conference on Computer Vision and Pattern
  Recognition, 2022, pp. 9924--9935.

\bibitem{dundar2020domain}
A.~Dundar, M.-Y. Liu, Z.~Yu, T.-C. Wang, J.~Zedlewski, J.~Kautz, Domain
  stylization: A fast covariance matching framework towards domain adaptation,
  IEEE Transactions on Pattern Analysis and Machine Intelligence 43~(7) (2020)
  2360--2372.

\bibitem{yang2020fda}
Y.~Yang, S.~Soatto, Fda: Fourier domain adaptation for semantic segmentation,
  in: Proceedings of the IEEE/CVF Conference on Computer Vision and Pattern
  Recognition, 2020, pp. 4085--4095.

\bibitem{hoffman2018cycada}
J.~Hoffman, E.~Tzeng, T.~Park, J.-Y. Zhu, P.~Isola, K.~Saenko, A.~Efros,
  T.~Darrell, Cycada: Cycle-consistent adversarial domain adaptation, in:
  International Conference on Machine Learning, Pmlr, 2018, pp. 1989--1998.

\bibitem{tsai2018learning}
Y.-H. Tsai, W.-C. Hung, S.~Schulter, K.~Sohn, M.-H. Yang, M.~Chandraker,
  Learning to adapt structured output space for semantic segmentation, in:
  Proceedings of the IEEE/CVF Conference on Computer Vision and Pattern
  Recognition, 2018, pp. 7472--7481.

\bibitem{vu2019advent}
T.-H. Vu, H.~Jain, M.~Bucher, M.~Cord, P.~P{\'e}rez, Advent: Adversarial
  entropy minimization for domain adaptation in semantic segmentation, in:
  Proceedings of the IEEE/CVF Conference on Computer Vision and Pattern
  Recognition, 2019, pp. 2517--2526.

\bibitem{zhang2021prototypical}
P.~Zhang, B.~Zhang, T.~Zhang, D.~Chen, Y.~Wang, F.~Wen, Prototypical pseudo
  label denoising and target structure learning for domain adaptive semantic
  segmentation, in: Proceedings of the IEEE/CVF Conference on Computer Vision
  and Pattern Recognition, 2021, pp. 12414--12424.

\bibitem{araslanov2021self}
N.~Araslanov, S.~Roth, Self-supervised augmentation consistency for adapting
  semantic segmentation, in: Proceedings of the IEEE/CVF Conference on Computer
  Vision and Pattern Recognition, 2021, pp. 15384--15394.

\bibitem{wu2021dannet}
X.~Wu, Z.~Wu, H.~Guo, L.~Ju, S.~Wang, Dannet: A one-stage domain adaptation
  network for unsupervised nighttime semantic segmentation, in: Proceedings of
  the IEEE/CVF Conference on Computer Vision and Pattern Recognition, 2021, pp.
  15769--15778.

\bibitem{sakaridis2019guided}
C.~Sakaridis, D.~Dai, L.~V. Gool, Guided curriculum model adaptation and
  uncertainty-aware evaluation for semantic nighttime image segmentation, in:
  Proceedings of the IEEE/CVF International Conference on Computer Vision,
  2019, pp. 7374--7383.

\bibitem{wang2020classes}
H.~Wang, T.~Shen, W.~Zhang, L.-Y. Duan, T.~Mei, Classes matter: A fine-grained
  adversarial approach to cross-domain semantic segmentation, in: Proceedings
  of the European Conference on Computer Vision, Springer, 2020, pp. 642--659.

\bibitem{li2019bidirectional}
Y.~Li, L.~Yuan, N.~Vasconcelos, Bidirectional learning for domain adaptation of
  semantic segmentation, in: Proceedings of the IEEE/CVF Conference on Computer
  Vision and Pattern Recognition, 2019, pp. 6936--6945.

\bibitem{li2023contrast}
T.~Li, S.~Roy, H.~Zhou, H.~Lu, S.~Lathuili{\`e}re, Contrast, stylize and adapt:
  Unsupervised contrastive learning framework for domain adaptive semantic
  segmentation, in: Proceedings of the IEEE/CVF Conference on Computer Vision
  and Pattern Recognition, 2023, pp. 4868--4878.

\bibitem{vu2019dada}
T.-H. Vu, H.~Jain, M.~Bucher, M.~Cord, P.~P{\'e}rez, Dada: Depth-aware domain
  adaptation in semantic segmentation, in: Proceedings of the IEEE/CVF
  International Conference on Computer Vision, 2019, pp. 7364--7373.

\bibitem{wang2021domain}
Q.~Wang, D.~Dai, L.~Hoyer, L.~Van~Gool, O.~Fink, Domain adaptive semantic
  segmentation with self-supervised depth estimation, in: Proceedings of the
  IEEE/CVF International Conference on Computer Vision, 2021, pp. 8515--8525.

\bibitem{xie2023sepico}
B.~Xie, S.~Li, M.~Li, C.~H. Liu, G.~Huang, G.~Wang, Sepico: Semantic-guided
  pixel contrast for domain adaptive semantic segmentation, IEEE Transactions
  on Pattern Analysis and Machine Intelligence (2023).

\bibitem{hoffman2016fcns}
J.~Hoffman, D.~Wang, F.~Yu, T.~Darrell, Fcns in the wild: Pixel-level
  adversarial and constraint-based adaptation, arXiv preprint arXiv:1612.02649
  (2016).

\bibitem{peng2023diffusion}
D.~Peng, P.~Hu, Q.~Ke, J.~Liu, Diffusion-based image translation with label
  guidance for domain adaptive semantic segmentation, in: Proceedings of the
  IEEE/CVF International Conference on Computer Vision, 2023, pp. 808--820.

\bibitem{saito2018maximum}
K.~Saito, K.~Watanabe, Y.~Ushiku, T.~Harada, Maximum classifier discrepancy for
  unsupervised domain adaptation, in: Proceedings of the IEEE/CVF Conference on
  Computer Vision and Pattern Recognition, 2018, pp. 3723--3732.

\bibitem{mei2020instance}
K.~Mei, C.~Zhu, J.~Zou, S.~Zhang, Instance adaptive self-training for
  unsupervised domain adaptation, in: Proceedings of the European Conference on
  Computer Vision, Springer, 2020, pp. 415--430.

\bibitem{zhang2018collaborative}
W.~Zhang, W.~Ouyang, W.~Li, D.~Xu, Collaborative and adversarial network for
  unsupervised domain adaptation, in: Proceedings of the IEEE/CVF Conference on
  Computer Vision and Pattern Recognition, 2018, pp. 3801--3809.

\bibitem{pan2019transferrable}
Y.~Pan, T.~Yao, Y.~Li, Y.~Wang, C.-W. Ngo, T.~Mei, Transferrable prototypical
  networks for unsupervised domain adaptation, in: Proceedings of the IEEE/CVF
  Conference on Computer Vision and Pattern Recognition, 2019, pp. 2239--2247.

\bibitem{zhou2023adaptive}
L.~Zhou, S.~Xiao, M.~Ye, X.~Zhu, S.~Li, Adaptive mutual learning for
  unsupervised domain adaptation, IEEE Transactions on Circuits and Systems for
  Video Technology (2023).

\bibitem{tranheden2021dacs}
W.~Tranheden, V.~Olsson, J.~Pinto, L.~Svensson, Dacs: Domain adaptation via
  cross-domain mixed sampling, in: Proceedings of the IEEE/CVF Winter
  Conference on Applications of Computer Vision, 2021, pp. 1379--1389.

\bibitem{li2022class}
R.~Li, S.~Li, C.~He, Y.~Zhang, X.~Jia, L.~Zhang, Class-balanced pixel-level
  self-labeling for domain adaptive semantic segmentation, in: Proceedings of
  the IEEE/CVF Conference on Computer Vision and Pattern Recognition, 2022, pp.
  11593--11603.

\bibitem{hoyer2022hrda}
L.~Hoyer, D.~Dai, L.~Van~Gool, Hrda: Context-aware high-resolution
  domain-adaptive semantic segmentation, in: Proceedings of the European
  Conference on Computer Vision, Springer, 2022, pp. 372--391.

\bibitem{hoyer2023mic}
L.~Hoyer, D.~Dai, H.~Wang, L.~Van~Gool, Mic: Masked image consistency for
  context-enhanced domain adaptation, in: Proceedings of the IEEE/CVF
  Conference on Computer Vision and Pattern Recognition, 2023, pp.
  11721--11732.

\bibitem{vilnis2014word}
L.~Vilnis, A.~McCallum, Word representations via gaussian embedding, arXiv
  preprint arXiv:1412.6623 (2014).

\bibitem{kingma2013auto}
D.~P. Kingma, M.~Welling, Auto-encoding variational bayes, arXiv preprint
  arXiv:1312.6114 (2013).

\bibitem{oh2018modeling}
S.~J. Oh, K.~Murphy, J.~Pan, J.~Roth, F.~Schroff, A.~Gallagher, Modeling
  uncertainty with hedged instance embedding, arXiv preprint arXiv:1810.00319
  (2018).

\bibitem{li2021spherical}
S.~Li, J.~Xu, X.~Xu, P.~Shen, S.~Li, B.~Hooi, Spherical confidence learning for
  face recognition, in: Proceedings of the IEEE/CVF Conference on Computer
  Vision and Pattern Recognition, 2021, pp. 15629--15637.

\bibitem{kirchhof2022non}
M.~Kirchhof, K.~Roth, Z.~Akata, E.~Kasneci, A non-isotropic probabilistic take
  on proxy-based deep metric learning, in: Proceedings of the European
  Conference on Computer Vision, Springer, 2022, pp. 435--454.

\bibitem{shi2019probabilistic}
Y.~Shi, A.~K. Jain, Probabilistic face embeddings, in: Proceedings of the
  IEEE/CVF International Conference on Computer Vision, 2019, pp. 6902--6911.

\bibitem{chun2021probabilistic}
S.~Chun, S.~J. Oh, R.~S. De~Rezende, Y.~Kalantidis, D.~Larlus, Probabilistic
  embeddings for cross-modal retrieval, in: Proceedings of the IEEE/CVF
  Conference on Computer Vision and Pattern Recognition, 2021, pp. 8415--8424.

\bibitem{neculai2022probabilistic}
A.~Neculai, Y.~Chen, Z.~Akata, Probabilistic compositional embeddings for
  multimodal image retrieval, in: Proceedings of the IEEE/CVF Conference on
  Computer Vision and Pattern Recognition, 2022, pp. 4547--4557.

\bibitem{xie2023boosting}
H.~Xie, C.~Wang, M.~Zheng, M.~Dong, S.~You, C.~Fu, C.~Xu, Boosting
  semi-supervised semantic segmentation with probabilistic representations, in:
  Proceedings of the AAAI Conference on Artificial Intelligence, Vol.~37, 2023,
  pp. 2938--2946.

\bibitem{hadsell2006dimensionality}
R.~Hadsell, S.~Chopra, Y.~LeCun, Dimensionality reduction by learning an
  invariant mapping, in: Proceedings of the IEEE/CVF Conference on Computer
  Vision and Pattern Recognition, Vol.~2, 2006, pp. 1735--1742.

\bibitem{he2020momentum}
K.~He, H.~Fan, Y.~Wu, S.~Xie, R.~Girshick, Momentum contrast for unsupervised
  visual representation learning, in: Proceedings of the IEEE/CVF Conference on
  Computer Vision and Pattern Recognition, 2020, pp. 9729--9738.

\bibitem{huang2022category}
J.~Huang, D.~Guan, A.~Xiao, S.~Lu, L.~Shao, Category contrast for unsupervised
  domain adaptation in visual tasks, in: Proceedings of the IEEE/CVF Conference
  on Computer Vision and Pattern Recognition, 2022, pp. 1203--1214.

\bibitem{jiang2022prototypical}
Z.~Jiang, Y.~Li, C.~Yang, P.~Gao, Y.~Wang, Y.~Tai, C.~Wang, Prototypical
  contrast adaptation for domain adaptive semantic segmentation, in:
  Proceedings of the European Conference on Computer Vision, Springer, 2022,
  pp. 36--54.

\bibitem{scott2021mises}
T.~R. Scott, A.~C. Gallagher, M.~C. Mozer, von mises-fisher loss: An
  exploration of embedding geometries for supervised learning, in: Proceedings
  of the IEEE/CVF International Conference on Computer Vision, 2021, pp.
  10612--10622.

\bibitem{richter2016playing}
S.~R. Richter, V.~Vineet, S.~Roth, V.~Koltun, Playing for data: Ground truth
  from computer games, in: Proceedings of the European Conference on Computer
  Vision, Springer, 2016, pp. 102--118.

\bibitem{ros2016synthia}
G.~Ros, L.~Sellart, J.~Materzynska, D.~Vazquez, A.~M. Lopez, The synthia
  dataset: A large collection of synthetic images for semantic segmentation of
  urban scenes, in: Proceedings of the IEEE/CVF Conference on Computer Vision
  and Pattern Recognition, 2016, pp. 3234--3243.

\bibitem{yu2020bdd100k}
F.~Yu, H.~Chen, X.~Wang, W.~Xian, Y.~Chen, F.~Liu, V.~Madhavan, T.~Darrell,
  Bdd100k: A diverse driving dataset for heterogeneous multitask learning, in:
  Proceedings of the IEEE/CVF Conference on Computer Vision and Pattern
  Recognition, 2020, pp. 2636--2645.

\bibitem{dai2018dark}
D.~Dai, L.~Van~Gool, Dark model adaptation: Semantic image segmentation from
  daytime to nighttime, in: 2018 21st International Conference on Intelligent
  Transportation Systems (ITSC), IEEE, 2018, pp. 3819--3824.

\bibitem{cheng2023adpl}
Y.~Cheng, F.~Wei, J.~Bao, D.~Chen, W.~Zhang, Adpl: Adaptive dual path learning
  for domain adaptation of semantic segmentation, IEEE Transactions on Pattern
  Analysis and Machine Intelligence (2023).

\bibitem{ren2023prototypical}
Q.~Ren, Q.~Mao, S.~Lu, Prototypical bidirectional adaptation and learning for
  cross-domain semantic segmentation, IEEE Transactions on Multimedia (2023).

\bibitem{zhao2023learning}
D.~Zhao, S.~Wang, Q.~Zang, D.~Quan, X.~Ye, R.~Yang, L.~Jiao, Learning
  pseudo-relations for cross-domain semantic segmentation, in: Proceedings of
  the IEEE/CVF International Conference on Computer Vision, 2023, pp.
  19191--19203.

\bibitem{truong2023fredom}
T.-D. Truong, N.~Le, B.~Raj, J.~Cothren, K.~Luu, Fredom: Fairness domain
  adaptation approach to semantic scene understanding, in: Proceedings of the
  IEEE/CVF Conference on Computer Vision and Pattern Recognition, 2023, pp.
  19988--19997.

\bibitem{shen2023diga}
F.~Shen, A.~Gurram, Z.~Liu, H.~Wang, A.~Knoll, Diga: Distil to generalize and
  then adapt for domain adaptive semantic segmentation, in: Proceedings of the
  IEEE/CVF Conference on Computer Vision and Pattern Recognition, 2023, pp.
  15866--15877.

\bibitem{fan2024otclda}
Q.~Fan, X.~Shen, S.~Ying, S.~Du, Otclda: Optimal transport and contrastive
  learning for domain adaptive semantic segmentation, IEEE Transactions on
  Intelligent Transportation Systems (2024).

\bibitem{lai2022decouplenet}
X.~Lai, Z.~Tian, X.~Xu, Y.~Chen, S.~Liu, H.~Zhao, L.~Wang, J.~Jia, Decouplenet:
  Decoupled network for domain adaptive semantic segmentation, in: Proceedings
  of the European Conference on Computer Visionn, Springer, 2022, pp. 369--387.

\bibitem{liu2023improving}
W.~Liu, W.~Li, J.~Zhu, M.~Cui, X.~Xie, L.~Zhang, Improving nighttime
  driving-scene segmentation via dual image-adaptive learnable filters, IEEE
  Transactions on Circuits and Systems for Video Technology (2023).

\bibitem{shen2023loopda}
F.~Shen, Z.~Pataki, A.~Gurram, Z.~Liu, H.~Wang, A.~Knoll, Loopda: Constructing
  self-loops to adapt nighttime semantic segmentation, in: Proceedings of the
  IEEE/CVF Winter Conference on Applications of Computer Vision, 2023, pp.
  3256--3266.

\bibitem{wu2021one}
X.~Wu, Z.~Wu, L.~Ju, S.~Wang, A one-stage domain adaptation network with image
  alignment for unsupervised nighttime semantic segmentation, IEEE Transactions
  on Pattern Analysis and Machine Intelligence 45~(1) (2021) 58--72.

\bibitem{gao2022cross}
H.~Gao, J.~Guo, G.~Wang, Q.~Zhang, Cross-domain correlation distillation for
  unsupervised domain adaptation in nighttime semantic segmentation, in:
  Proceedings of the IEEE/CVF Conference on Computer Vision and Pattern
  Recognition, 2022, pp. 9913--9923.

\bibitem{sakaridis2020map}
C.~Sakaridis, D.~Dai, L.~Van~Gool, Map-guided curriculum domain adaptation and
  uncertainty-aware evaluation for semantic nighttime image segmentation, IEEE
  Transactions on Pattern Analysis and Machine Intelligence 44~(6) (2020)
  3139--3153.

\end{thebibliography}
\end{document}